\documentclass[letterpaper, 10 pt, journal, twoside]{IEEEtran}

\newif\ifRALfinal
\RALfinalfalse

\newif\ifpreprint
\ifRALfinal
  \preprintfalse
\else
  \preprinttrue
\fi

\IEEEoverridecommandlockouts                              

\makeatletter
    \let\NAT@parse\undefined
\makeatother

\usepackage{mathptmx} 
\usepackage{times} 
\usepackage{amsmath} 
\usepackage{amssymb}  
\usepackage{xcolor}
\usepackage{comment}
\usepackage{graphicx}
\usepackage{siunitx}
\usepackage{booktabs}
\usepackage{wrapfig}
\usepackage{adjustbox}
\usepackage{balance}
\usepackage[colorlinks = true,
            linkcolor = {red!50!black},
            urlcolor  = blue,
            citecolor = magenta,
            bookmarks = true,
            pagebackref]{hyperref}
\usepackage{cite}
\usepackage{subcaption}
\usepackage{caption}

\title{Data Scaling for Navigation in Unknown Environments}

\author{%
Lauri Suomela\(^{1}\), Naoki Takahata\(^{2}\), Sasanka Kuruppu Arachchige\(^{1}\), Harry Edelman\(^{3}\), Joni-Kristian Kämäräinen\(^{1}\)%
\ifRALfinal
\thanks{Manuscript received: January, 12, 2026; Accepted March, 11, 2026.}%
\thanks{This paper was recommended for publication by Editor J. Civera upon evaluation of the Reviewers’ comments.}%
\fi
\thanks{%
This work was supported by funding from
the European Union’s Horizon 2020 research and innovation programme under grant agreement No. 101034366,
and
the European Commission’s HORIZON.1.2 - Marie Skłodowska-Curie Actions (MSCA) under grant agreement No. 101072634.%
}
\thanks{\(^{1} \)Lauri Suomela, Sasanka Kuruppu Arachchige and Joni-Kristian Kämäräinen are with the Signal Processing Research Centre, Faculty of Information Technology and Communication Sciences, Tampere University, 33720 Tampere, Finland.
        Correspondence: {\tt\footnotesize lauriasuomela@gmail.com}}%
\thanks{\(^{2} \)Naoki Takahata is with the Graduate School of Information Sciences, Tohoku University, Sendai, Miyagi 980-8577, Japan.}%
\thanks{\(^{3} \)Harry Edelman is with the Faculty of Engineering, Turku University of Applied Sciences, 20520 Turku, Finland.}%
\ifRALfinal
\thanks{Digital Object Identifier (DOI): see top of this page.}
\fi
}

\makeatletter
\let\@oldmaketitle\@maketitle%
\renewcommand{\@maketitle}{\@oldmaketitle%
    \centering
    \vspace{0.15cm}
    \includegraphics[trim={0 0 0 1.4cm},clip,width=\linewidth]{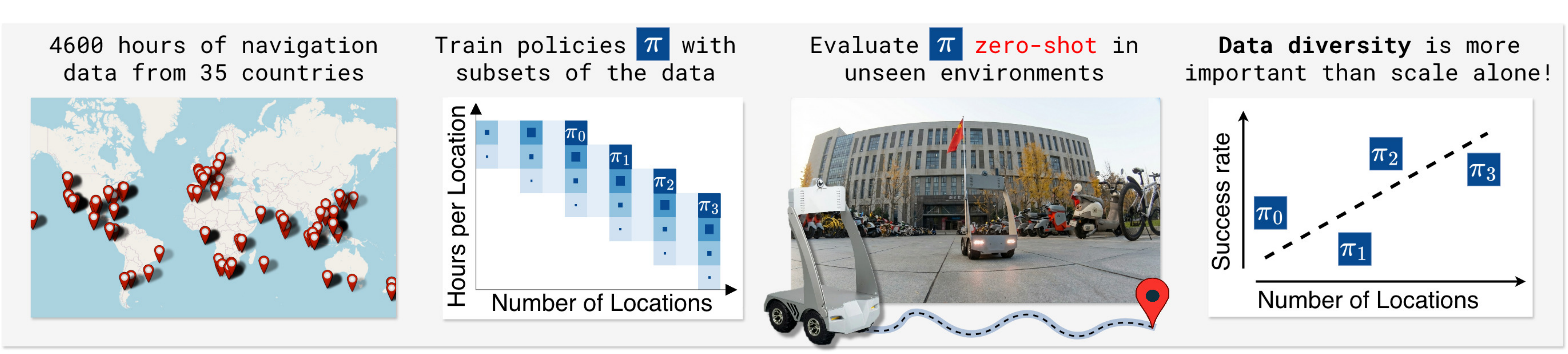}
    \captionof{figure}{We present an in-depth investigation of data scaling for end-to-end learned navigation.
    Real-world experiments demonstrate that generalization to unseen environments scales well with geographical diversity of training data, while benefit from data quantity alone saturates quickly.}
    \vspace{-0.8cm}
    \label{fig:pull_fig}
}
\addtocounter{figure}{-2}

\begin{document}
\makeatletter
\DeclareRobustCommand\onedot{\futurelet\@let@token\@onedot}
\def\@onedot{\ifx\@let@token.\else.\null\fi}

\def\eg{\emph{e.g}\onedot} \def\Eg{\emph{E.g}\onedot}
\def\ie{\emph{i.e}\onedot} \def\Ie{\emph{I.e}\onedot}
\def\cf{\emph{c.f}\onedot} \def\Cf{\emph{C.f}\onedot}
\def\etc{\emph{etc}\onedot} \def\vs{\emph{vs}\onedot}
\def\wrt{w.r.t\onedot} \def\dof{d.o.f\onedot}
\def\etal{et al\onedot}

\makeatother

\ifRALfinal
\markboth{IEEE Robotics and Automation Letters. Preprint Version. Accepted March, 2026}{Suomela \MakeLowercase{\textit{et al.}}: Data scaling for navigation in unknown environments}
\else
\markboth{}{}
\fi

\maketitle

\begin{abstract}
Generalization of imitation-learned navigation policies to environments unseen in training remains a major challenge. We address this by conducting the first large-scale study of how data quantity and data diversity affect real-world generalization in end-to-end, map-free visual navigation. Using a curated 4,565-hour crowd-sourced dataset collected across 161 locations in 35 countries, we train policies for point goal navigation and evaluate their closed-loop control performance on sidewalk robots operating in four countries, covering 125 km of autonomous driving.
Our results show that large-scale training data enables zero-shot navigation in unknown environments, approaching the performance of policies trained with environment-specific demonstrations. Critically, we find that data diversity is far more important than data quantity. Doubling the number of geographical locations in a training set decreases navigation errors by~$\sim15\%$, while performance benefit from adding data from existing locations saturates with very little data. We also observe that, with noisy crowd-sourced data, simple regression-based models outperform generative and sequence-based architectures. We release our policies, evaluation setup and example videos on the project page, \href{https://lasuomela.github.io/navigation_scaling/}{lasuomela.github.io/navigation\_scaling}.
\end{abstract}


\section{INTRODUCTION}

\IEEEPARstart{I}{n} recent years, learning-based approaches to robot navigation have received increasing interest. They enable communicating with humans in natural language, utilizing semantic cues in the environment, and performing tasks whose solutions are difficult to engineer by hand.
A key challenge in learning-based methods is generalization beyond locations seen in the training data. To be useful, a navigation algorithm should be able to function in a novel, unseen environment without additional finetuning.

Data scaling~\cite{rosenfeld_scaling_2021} is a promising approach to addressing generalization in robot learning, yet its different dimensions remain under-explored.
Recent work in robotic manipulation~\cite{lin_data_2025, black_pi05_2025, etukuru_robot_2025} shows that imitation learning policies' out-of-domain performance is strongly influenced by training data \emph{diversity} rather than quantity alone. Whether similar scaling behavior holds for navigation is unclear, as navigation involves long-horizon, closed-loop decision making where errors compound over time.
We hypothesize that the geographic variability encountered during training is a major driver of diversity that benefits navigation performance. Different locations naturally vary in visual appearance, material composition, and route structures.

Most prior work on learning-based navigation has not explicitly examined how training data diversity impacts generalization, in part due to the lack of sufficiently large and geographically diverse real-world datasets. Collecting expert navigation demonstrations across many locations is labor-intensive.
In this work, we achieve sufficient scale by leveraging non-expert, crowd-sourced demonstrations collected by a globally distributed fleet of low-cost sidewalk robots teleoperated over the internet.

We present an analysis, illustrated in Fig.~\ref{fig:pull_fig}, that disentangles how training data quantity and geographic diversity individually contribute to navigation performance in unknown environments. Specifically, we focus on vision-based navigation methods that formulate navigation as an end-to-end imitation learning task,
and use the number of distinct locations in the training set as a proxy measure for diversity.
We train map-free point-goal~\cite{anderson_evaluation_2018} navigation policies using different subsets of a crowd-sourced dataset comprising \SI{4600}{\hour} of demonstrations collected across 35 countries. These subsets jointly vary the number of training locations and the amount of data sampled from each location.
The resulting policies' generalization to environments not seen during training is evaluated on robots situated in four different countries.
In total, our experiments comprise 125 kilometers of real-world autonomous driving, which took 120 hours to complete.

We find that a policy trained with diverse large-scale data can control a robot in an unseen environment better than an in-domain policy trained specifically for that environment, and almost as well as a policy incorporating both large-scale and in-domain data.
Analysis of the individual data scale factors shows that increasing the amount of training data from any given set of locations saturates with very little data per location. In contrast, the number of navigation failures consistently decreases with a power-law relationship to the number of distinct training locations.
Doubling the number of training locations leads to a \(\sim15\%\) drop in navigation failures.

\noindent
As a summary of our contributions,

\begin{itemize}

    \item We perform the first in-depth, real-world investigation of data scaling for learning-based autonomous navigation.

    \item We show that generalization to unseen environments scales well with increased geographical diversity of the training data, while benefit from additional data from existing locations saturates surprisingly quickly.

    \item To support reproducible research, we open-source the trained policies, all code and the deployment setup. The dataset is available upon request.
    
\end{itemize}

\section{RELATED WORK}
\noindent
\textbf{Vision-based Navigation in Unknown Environments.}
Navigation in unknown environments is commonly implemented as finding minimum-cost paths through cost maps with geometric~\cite{elfes_using_1989, fox_dynamic_1997, lu_layered_2014} or learned~\cite{frey_fast_2023, zhang_creste_2025, sivaprakasam_salon_2025} costs.
They often rely on depth measurements and accurate synchronization between different sensors.
This makes the methods challenging to implement on low-cost robot platforms with limited sensing.
In this work we focus on end-to-end imitation learning approaches that can be deployed with a minimal set of sensors.
Evaluating such policies in environments unseen during training is standard practice.
However, many of the works with real-world deployment~\cite{shah_viking_2022, shahViNTLargeScaleMultiTask2023, suomela_placenav_2024, shen_effonav_2025} utilize image-goals, and often require prior maps of the environment or only perform short-horizon navigation.
Point goal navigation~\cite{anderson_evaluation_2018} is an alternative form of goal specification that is better suited for long-horizon tasks.
Still, most works on the point goal navigation are limited to short-horizon goal reaching~\cite{liu_citywalker_2025} or simulation~\cite{wijmansDDPPOLearningNearPerfect2020}.
In this work we focus on the topic of long-horizon point goal navigation.
While the experiments in the existing work~\cite{kahn_badgr_2021, he_seeing_2025, hirose_learning_2025, hirose_omnivla_2025} tend to be centered around a single geographic location, we deploy policies in multiple different countries without any prior knowledge of the environment.

\vspace{0.2cm}
\noindent
\textbf{Data Scaling in Robotics.} Most work on scaling robotics datasets has been conducted in the context of robotic manipulation~\cite{sartor_neural_2025}. Recent studies~\cite{black_pi05_2025, etukuru_robot_2025, lin_data_2025, villasevil_robot_2025} suggest that for generalization to unseen environments, dataset \emph{diversity} from scaling the number of locations from which training data is sampled is more important than absolute data quantity.
The phenomenon, however, has not been thoroughly investigated in the context of navigation.
Several works present some data scaling experiments, but the results' significance is limited by use of open-loop metrics~\cite{zheng_preliminary_2024, liu_citywalker_2025} or experimentation in simulation~\cite{zheng_preliminary_2024, xie_vid2sim_2025}. Furthermore, they do not explicitly address generalization to novel environments or examine dataset composition.
In this work, we specifically investigate how policy performance in unseen environments depends on both the quantity and the geographic diversity of training data. Our analysis is based on the results of closed-loop experiments conducted on real robots.

\vspace{0.2cm}
\noindent
\textbf{Navigation Datasets.}
Similar to most areas of robotics, data for training vision-based navigation models is scarce. Commonly utilized datasets contain around \SI{100}{\hour} of demonstrations~\cite{shahGNMGeneralNavigation2023, hirose_lelan_2024} from a limited number of locations.
Increasing the amount of training data has been explored using several different approaches.
Simulation is an attractive option for generating large amounts of demonstrations~\cite{wijmansDDPPOLearningNearPerfect2020, ehsani_spoc_2024, suomela_synthetic_2025}, but models trained with synthetic data suffer from the simulation-to-real gap.
Autonomous driving datasets are larger, in the order of thousands of hours~\cite{xu_end--end_2017, noauthor_lerobot_2025}, but closed-loop evaluation on an autonomous vehicle is risky and deployment on smaller platforms incurs an embodiment gap.
Yet another avenue is to train with generic internet videos~\cite{lin_learning_2023, liu_citywalker_2025}, but the resulting policies need to be fine-tuned with embodiment-specific data before deployment.
Finally, embodiment-specific data collection can be scaled by crowd-sourcing the process~\cite{muller_openbot-fleet_2024, frodobots_frodobots-2k_nodate}.
Imitating these potentially sub-optimal demonstrations using noisy, low-cost sensors can be challenging~\cite{hirose_learning_2025}, but we find that rigorous data cleaning enables us to train navigation policies even from such data. 
In this work, we utilize a dataset extracted from~\SI{8,000}{\hour} of raw crowd-sourced demonstrations.


\section{METHODS}

\noindent
\textbf{Problem statement.}
We consider PointGoal~\cite{anderson_evaluation_2018} navigation, where an agent is tasked with navigating to goals specified as 2D coordinates without access to a map. At each time step \(t\), the agent receives an RGB image observation \(O_t\) and the (noisy) distance and direction
\(g_t =
\big[\begin{smallmatrix}
  d\\
  \theta
\end{smallmatrix}\big]\)
to the goal coordinate \(G\).
The navigation agent learns a policy
\begin{equation}
    \pi(\mathbf{a}_t | \mathbf{O}_{t}, \mathbf{g}_{t})
\end{equation}
that maps the sequence of \(P\) most recent observations \(\mathbf{O}_t = \left\{ O_{t-P+1}, \ldots, O_{t} \right\}\), \(\mathbf{g}_t = \left\{ g_{t-P+1}, \ldots, g_{t} \right\}\) to an action chunk \(\mathbf{a}_t = \left\{ a_t, \ldots, a_{t+H-1} \right\}\) of \(H\) future actions towards the goal.

\vspace{0.2cm}
\noindent
\textbf{FrodoBots8k Dataset.}
We train the policies on navigation data collected using the FrodoBots AI crowd-sourcing platform~\cite{frodobots_frodobots_nodate}.
To incentivize the general public to contribute data, the platform casts the task as a game where players teleoperate actual robots over the internet.
The robots are located across different countries around the world. Since each robot only produces data from the immediate surroundings of its 'home base', the resulting dataset is geographically clustered into distinct locations. For each location, there are multiple repetitions of roughly the same route, performed at distinct times and dates by different robot operators.

We curate our training dataset from a larger batch of raw teleoperation data that totals around \SI{8000}{\hour} of navigation.
To cluster the data geographically, we first find pairs of navigation episodes whose GPS locations are less than \SI{100}{\meter} apart. Then, the pair-wise connectivity is propagated to find connected sets that we refer to as distinct training data locations.
To improve data quality, we remove navigation episodes and episode segments where the robot is stationary or any sensor measurements are missing. After filtering, we are left with \SI{4565}{\hour} of navigation from 161 distinct locations across 35 different countries on all continents except Antarctica.

Since the robots’ low-quality sensors are prone to noise, we improve pose estimates via sensor fusion. We combine GPS measurements, magnetometer readings, and wheel RPMs using non-linear factor graph optimization with GTSAM~\cite{frank_dellaert_and_gtsam_contributors_borglabgtsam_2022}. However, faulty magnetometers on some robots limit the accuracy of the resulting orientation estimates.

\begin{figure}[t]
    \centering
    \smallskip
    \smallskip
    \includegraphics[width=1.0\linewidth]{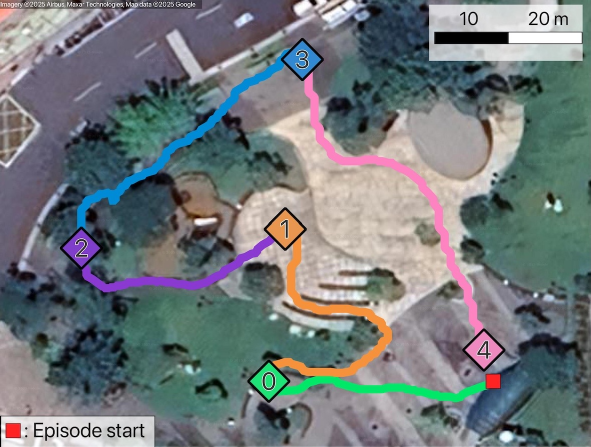}
    \caption{A crowd-sourced navigation episode segmented into 5 demonstrations for policy training.}
    \label{fig:goal_segmentation}
    \vspace{-0.5cm}
\end{figure}

The recorded navigation episodes do not contain information about which goal coordinate the player was trying to reach at any given time, and the routes are often circular. To sample plausible demonstrations of reaching some goal, we divide the navigation episodes into segments 
by finding peaks in the relationship between traveled path length and the Euclidean distance from segment start location. An example is shown in Fig.~\ref{fig:goal_segmentation}.

Finally, all sensors are subsampled and aligned to \SI{4}{\hertz}. The player-issued robot velocity commands are resampled so that each sensor time step \(t\) receives an associated \SI{10}{\hertz} chunk \(\mathbf{a}_{t}^{gt}\) of actions over the next \SI{1}{\second}.
Any demonstrations from locations closer than \SI{95}{\kilo\meter} to our test sites are removed from the training set.

\vspace{0.2cm}
\noindent
\textbf{Policy design.}
We experiment with multiple alternative policy architectures that all share the same basic structure.
Image observations \(O_t\) are encoded by a vision transformer into patch tokens that are compressed into a single feature vector with a 2D convolutional layer, similar to~\cite{majumdar_where_2024}.
The distances and directions to goal \(g_t\) are encoded into a \(64D\) vector with a 3-layer MLP followed by LayerNorm.
From here on, the different architectures diverge (see Sec.~\ref{sec:exp_ablation}), but ultimately all of them produce an action chunk \(\mathbf{a}_t\) of robot control commands.
Based on the model comparison study in Sec.~\ref{sec:exp_ablation}, we chose to perform our main experiments with the \textbf{MLP-BC} policy using the Theia-Base~\cite{shang_theia_2024} vision encoder. The MLP-BC policy simply concatenates the latest (\(P=1\)) image embedding and goal embedding, and an MLP head maps the concatenated vector into an action chunk.

\vspace{0.2cm}
\noindent
\textbf{Training details.}
The policies are trained to imitate the player-issued velocity-space robot control commands given the associated observations.
A training sample consists of \(P\) front camera observations \(\mathbf{O}_t\), and the distances and directions \(\mathbf{g}_t\) to a goal location \(G\) for each observation.
\(\mathbf{O}_t\) is picked from a demonstration trajectory at a random time \(t\), and \(G\) is sampled from the robot's future trajectory within \SI{150}{\meter} traveled distance from the robot location at observation time \(t\).

The images are resized to \(224 \times 224\).
The goal inputs \(g_t\) are obtained as the Euclidean distance and direction to \(G\) from the robot's pose at time \(t\). The goal direction \(\theta \in [-\pi, \pi]\) is normalized to \([-1, 1]\), and the distance \(d\) is clipped to \([0, 1]\) in kilometers.
The 2D action-space 
\(
\mathcal{A} \subset \mathbb{R}^2,~ 
a = (v, \omega)
\)
of forward linear and yaw angular velocities is normalized to \([-1, 1]\). The chunks \(\mathbf{a}_{t}^{gt}\) of \(H=10\) actions issued by a player over \SI{1}{\second} window are used as prediction targets.

We train all policies with batch size of 2048, an AdamW optimizer and a learning rate of \(1e-4\) decayed with cosine schedule. Training is distributed across 16 AMD MI250x GPUs. The policies with a generative objective are trained with the L2 loss.
Policies trained for a regression objective with the regular L2 loss, however, learn to just drive straight forward unless about to collide with a obstacle.
Given that a major part of the dataset demonstrations consists of driving straight ahead, we observe better goal-following behavior when we increase the weight of the L2 loss \(\mathcal{L}_2\) for samples with non-zero target angular velocity as

\begin{equation}
    \mathcal{L}_{scaled} = \mathcal{L}_2 \cdot(s_{min} + (s_{max} - s_{min})\cdot|\omega|),
\end{equation}

where \(s_{min}\) and \(s_{max}\) are the minimum and maximum scaling weights, and \(|\omega| \in [0,1]\) is the largest angular velocity magnitude in the associated \(\mathbf{a}_{t}^{gt}\). We set \(s_{min}\) to 1 and \(s_{max}\) to 10.
We apply standard color jitter augmentation to the images. The \(\mathbf{O}_t\), \(\mathbf{g}_t\) and \(\mathbf{a}_{t}^{gt}\) are mirrored horizontally with 50\% probability. We train all policies for two epochs, the exact number of optimizer steps depending on dataset size.

\section{EXPERIMENTS}
\begin{figure*}[t]
\smallskip
\smallskip
  \centering
  \begin{subfigure}[b]{0.245\textwidth}
    \includegraphics[width=\linewidth]{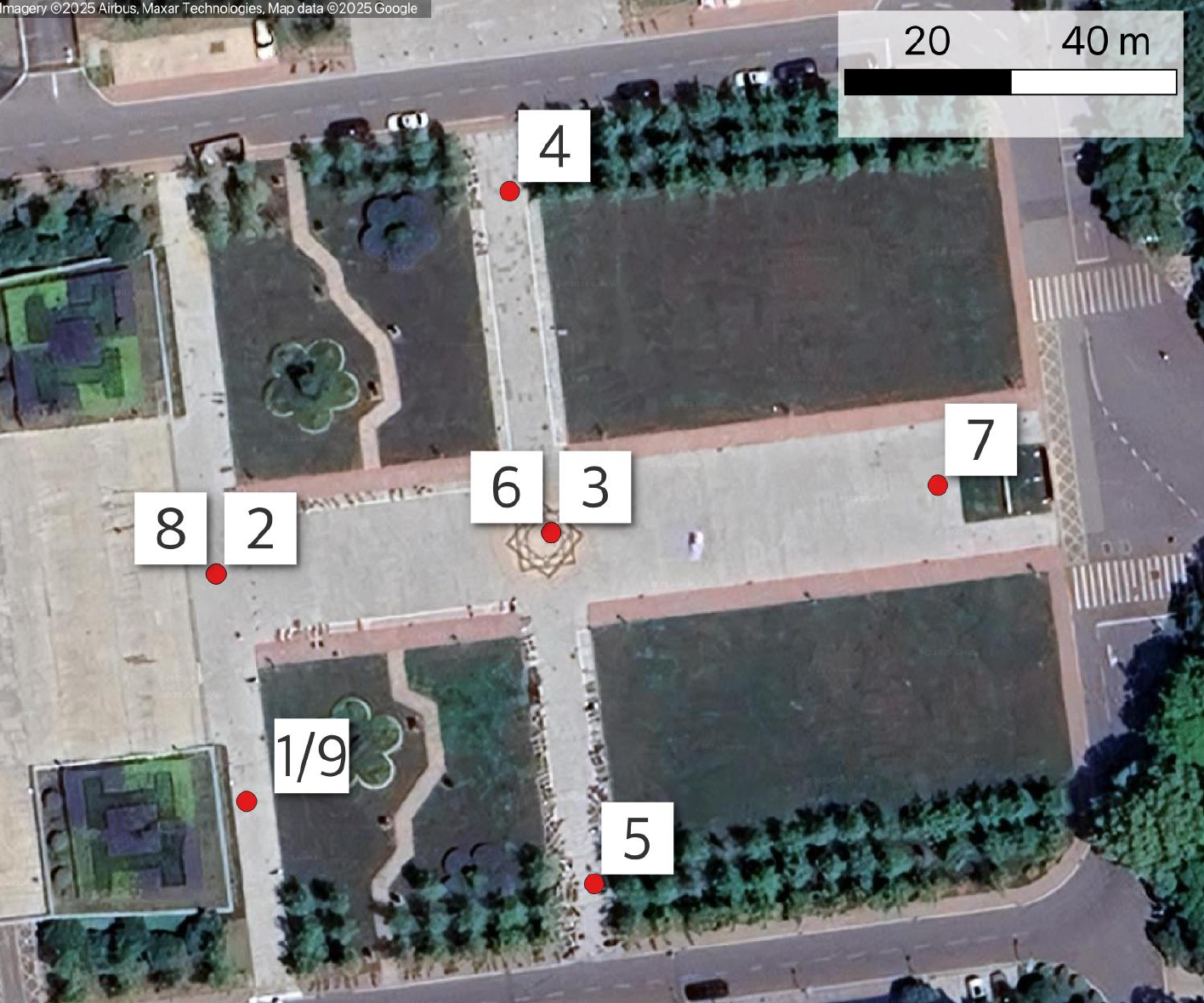}
    \vspace{0.5em}
    \includegraphics[width=\linewidth]{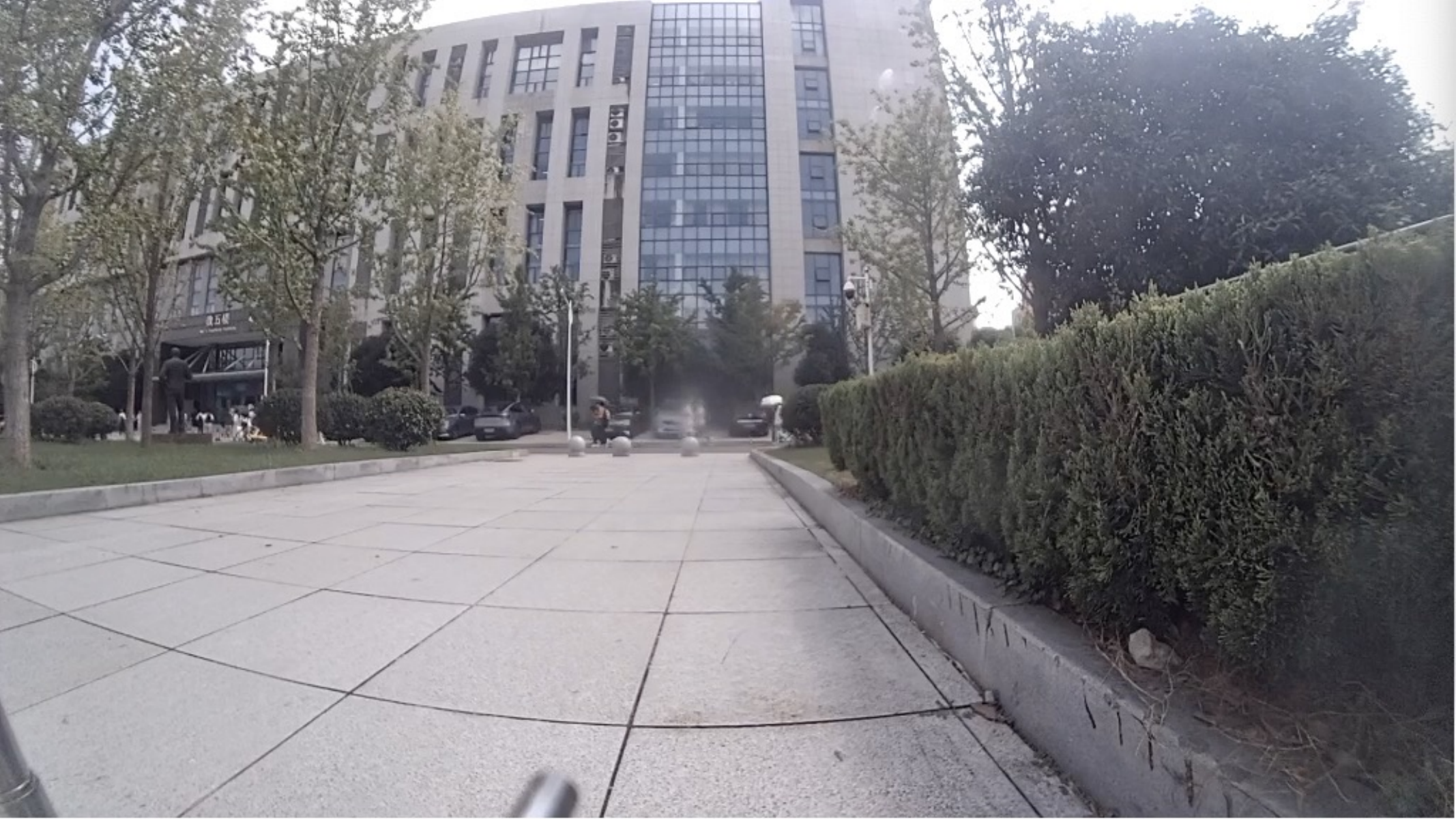}
    \caption{
        \raggedright Wuhan, China.\linebreak{\scriptsize\texttt{
        WGS84 (30.48244, 114.30264)
        }}
    } 
    \label{fig:wuhan}
  \end{subfigure}
  \hfill
  \begin{subfigure}[b]{0.245\textwidth}
    \includegraphics[width=\linewidth]{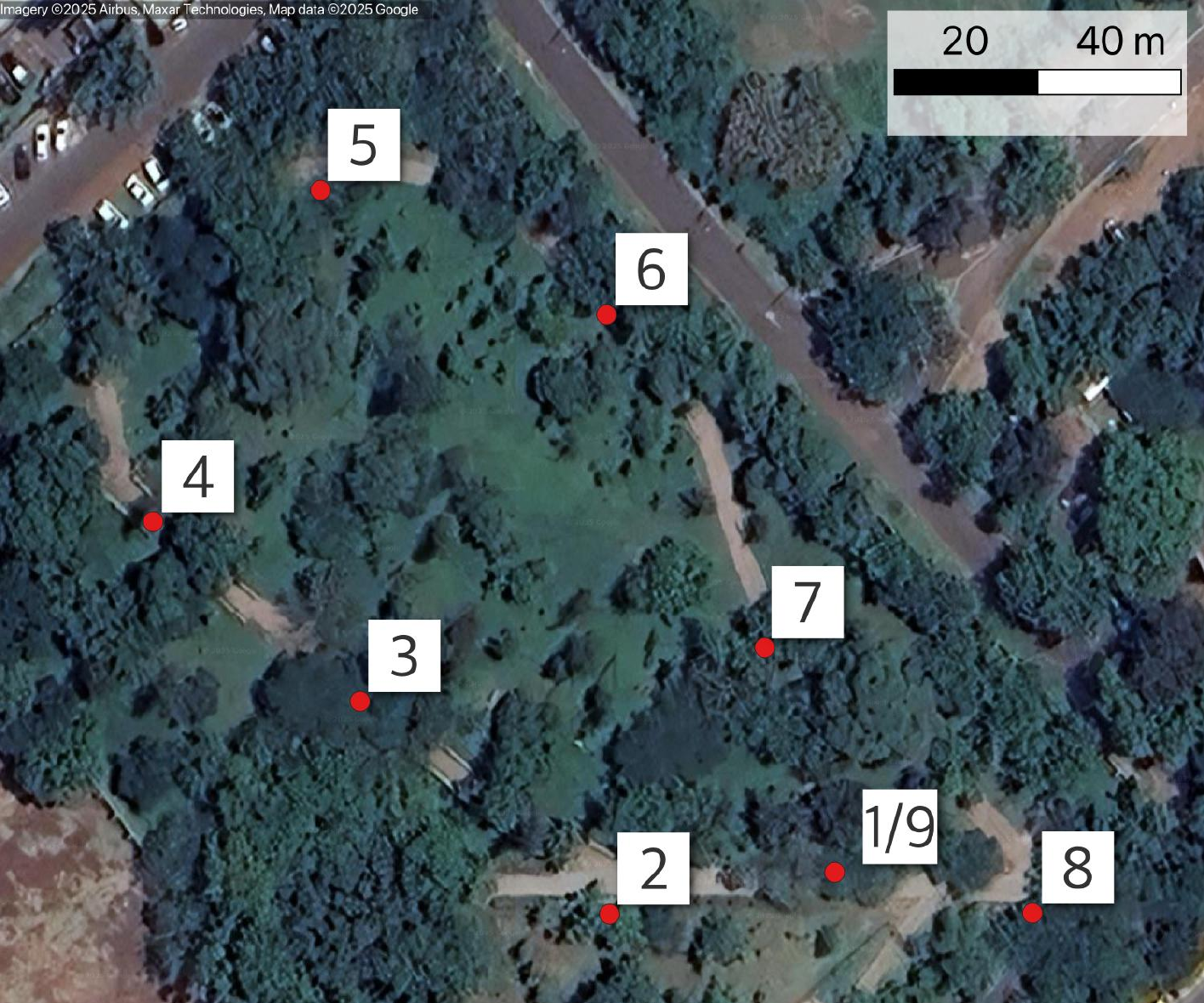}
    \vspace{0.5em}
    \includegraphics[width=\linewidth]{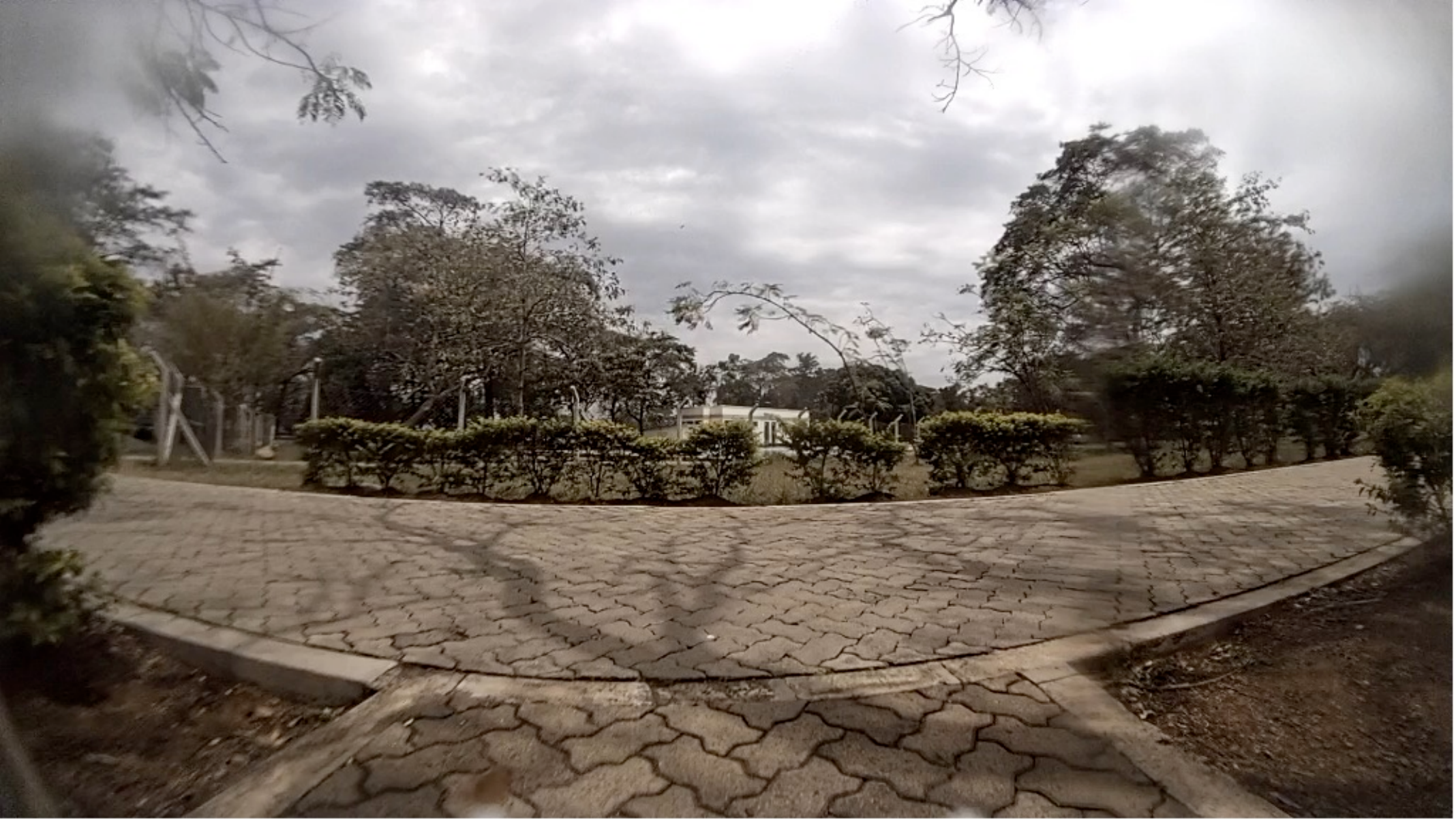}
    \caption{
        \raggedright Kisumu, Kenya.
        \linebreak{\scriptsize\texttt{
        (-0.11052, 34.75131)
        }}
    }
    \label{fig:kisumu}
  \end{subfigure}
  \hfill
  \begin{subfigure}[b]{0.245\textwidth}
    \includegraphics[width=\linewidth]{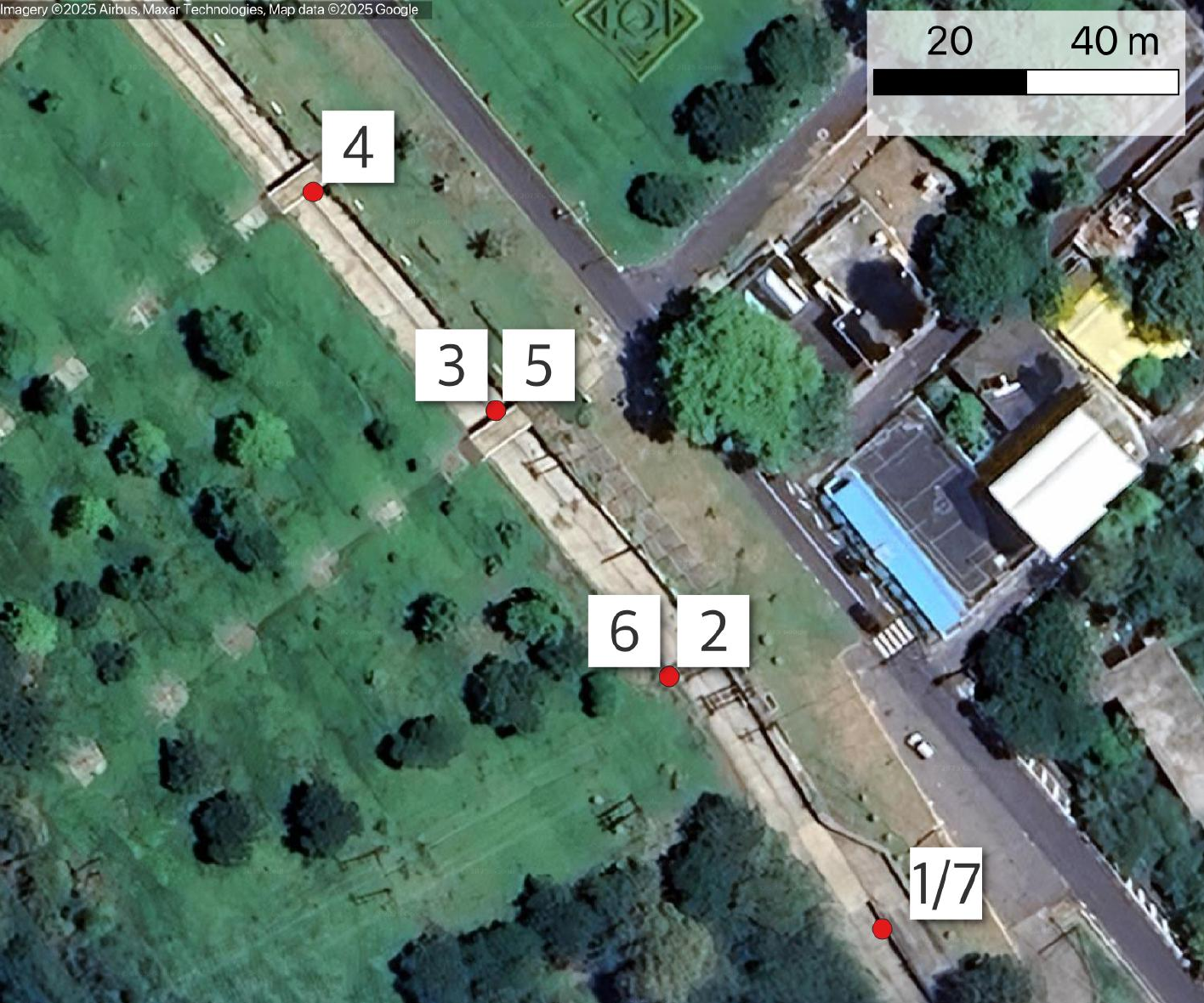}
    \vspace{0.5em}
    \includegraphics[width=\linewidth]{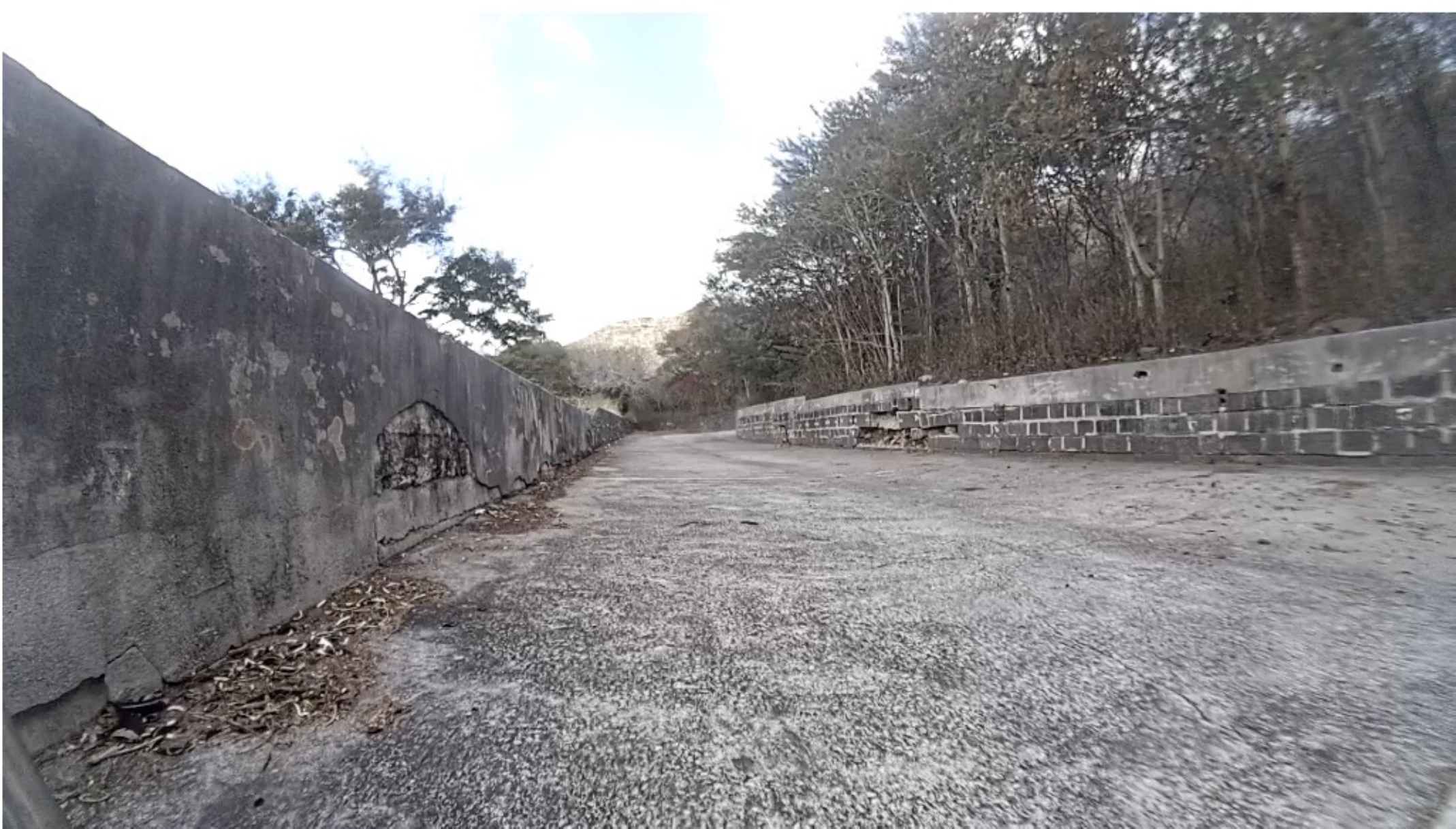}
    \caption{
        \raggedright Port Louis, Mauritius.
        \linebreak{\scriptsize\texttt{
        (-20.17148, 57.49782)
        }}
    }
    \label{fig:portlouis}
  \end{subfigure}
  \hfill
  \begin{subfigure}[b]{0.245\textwidth}
    \includegraphics[width=\linewidth]{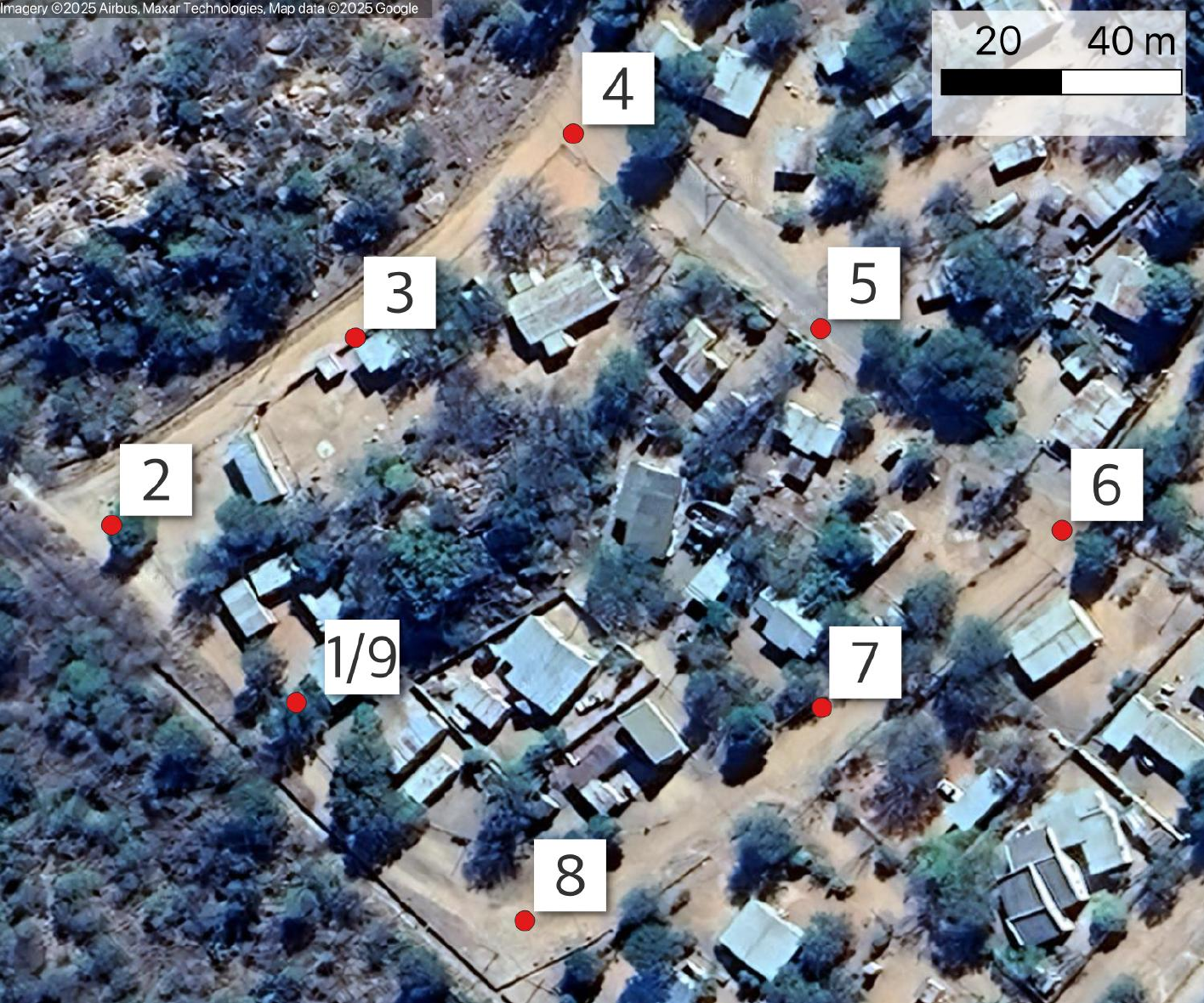}
    \vspace{0.5em}
    \includegraphics[width=\linewidth]{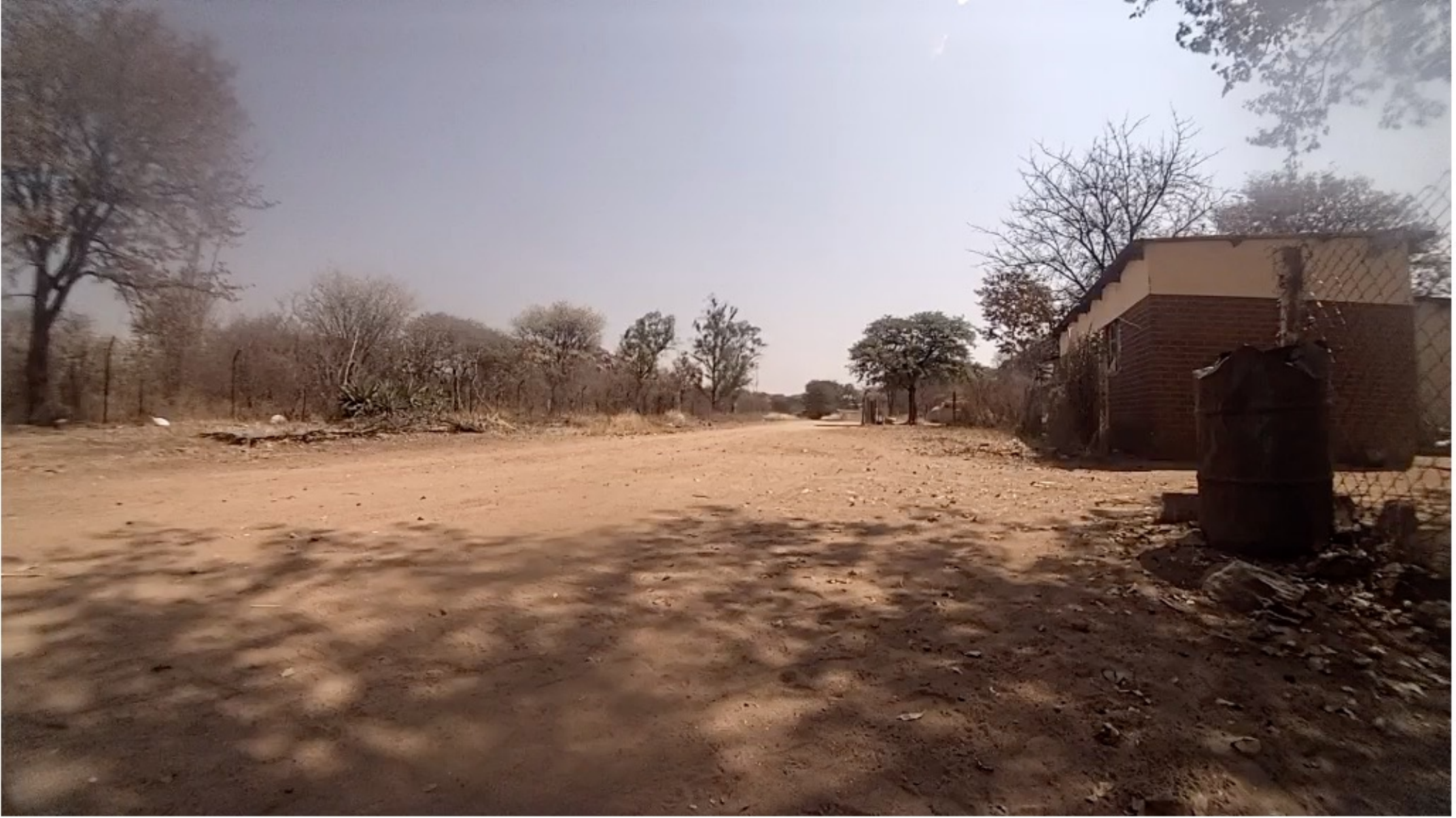}
    \caption{
        \raggedright Selebi Phikwe, Botswana.
        \linebreak{\scriptsize\texttt{
        (-21.98324, 27.83091)
        }}
    }
    \label{fig:selebiphikwe}
  \end{subfigure}

  \caption{The test route segment checkpoints and robot front camera views in the different environments.}
  \label{fig:test_envs}
  \vspace{-0.4cm}
\end{figure*}

We performed a series of experiments to answer the following research questions:

\begin{itemize}

    \item \textbf{Q1}:
    Does training on large-scale data enable a policy to generalize and successfully navigate in unseen environments across diverse geographic regions?

    \item \textbf{Q2}: How does the generalization performance depend on the \emph{quantity} and \emph{diversity} of training data?
    
    \item \textbf{Q3}: How well do different policy architectures perform in a setting with crowd-sourced training data and remote inference?

\end{itemize}

\subsection{Experiment setup.}

\noindent
\textbf{Policy Deployment.}
We perform all experiments on Earth Rover Zero robots, the same robot type used for collecting the training dataset. It is a differential drive robot designed for remote teleoperation over the internet. The robots, located in different countries around the world, stream sensor observations to a remote machine that runs policy inference and sends control commands back to the robot. We utilize a desktop computer with an Nvidia RTX3090 GPU. With good network conditions, the round-trip latency including policy inference is \(<\)\SI{1}{\second}. The policies receive monocular RGB image observations from the robot's \SI{110}{\degree}~\emph{FOV} frontal camera, distance to goal calculated from GPS, and direction to goal computed from magnetometer readings fused with optical flow by a Kalman filter. All policies run at \SI{4}{\hertz}.

\vspace{0.2cm}
\noindent
\textbf{Test environments.}
\label{sec:test_environments}
Policy performance is evaluated in 4 different locations situated in China, Kenya, Mauritius, and Botswana (Figure~\ref{fig:test_envs}). Each test route, ranging \SI{250}{\meter} to \SI{400}{\meter} in length, is divided into multiple \SI{30}{\meter}-\SI{90}{\meter} segments. To complete a full test route, the robot sequentially navigates the series of segment checkpoints. In total, our test environments contain 30 individual navigation segments.
Despite appearing as simple straight-line paths in Fig.~\ref{fig:test_envs}, in practice completing the segments requires~\eg~obstacle avoidance, terrain traversability estimation, and finding correct intersections. We refer the reader to the supplemental videos for illustration of task difficulty level.

\begin{figure*}[t]
    \centering
    \includegraphics[width=\textwidth]{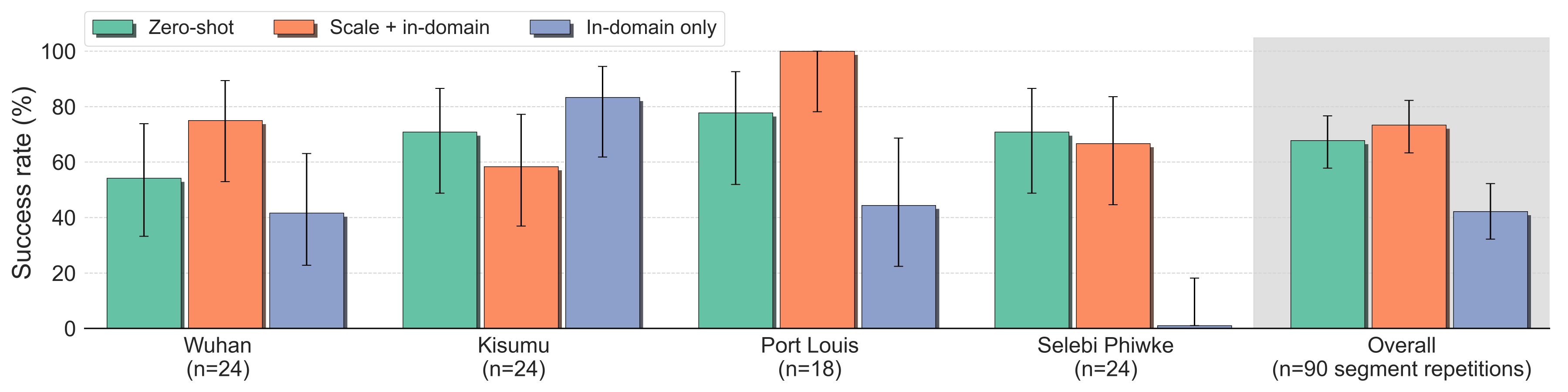}
    \caption{Policies trained with large-scale data achieve higher success rates compared to environment-specific policies. The zero-shot policy almost matches the policy that incorporates both large-scale data and data from the test environments.}
    \label{fig:generalization}
\end{figure*}

\vspace{0.2cm}
\noindent
\textbf{Evaluation and metrics.}
We use \emph{navigation success rate}~\cite{anderson_evaluation_2018}, the proportion of successful navigation attempts, as our main metric.
For increased granularity, we report success rates over \emph{segments} rather than evaluate route-level performance over the full routes.
If a policy fails a segment, we teleoperate the robot to the start location of the next segment and continue evaluation from there.
We consider a segment failed if the robot collides with an object and cannot independently get unstuck, deviates from the test route in such a way that there is no chance of recovering, or the robot operator has to intervene in order to prevent harm to the robot or bystanders. We do not count interventions needed because of external circumstances,~\eg~intentional harassment by humans or yielding to vehicles as safety precaution.
A segment is considered successful if the robot arrives within \SI{10}{\meter} of the segment checkpoint. The relatively loose success margin was chosen to accommodate GPS noise.
To capture variation in weather conditions, network latency and GPS drift, we test each policy multiple times in each environment on different dates and times of day.
However, as a single test run can take up to \SI{30}{\minute} and we test a large number of different policies, we needed to balance the number with practical feasibility. As compromise, we perform 3 repetitions of each route with each policy in all experiments.
We report the average success rate and its 95\% confidence interval from the continuity-corrected Wilson method~\cite{newcombe_two-sided_1998} over segments from the different test environments.
To support additional analysis, we provide auxiliary metrics for some of the experiments. We calculate
\emph{1) Normalized Intervention Rate (NIR)} - the number of interventions needed to make \SI{100}{\meter} of route progress
\emph{2) Normalized Progress Speed (NPS)} - duration (seconds) to make \SI{100}{\meter} of route progress, calculated from successful segments
\emph{3) Distance (Dist.)} - a route-level metric measuring progress along the test route until the first failure.

\subsection{Large-scale data enables navigation in unseen environments.}
\label{sec:exp_generalization}
To study \textbf{Q1.}, we perform an experiment to evaluate policy zero-shot navigation performance in unseen environments. To contextualize the result, we compare the zero-shot policy's performance with in-domain policies trained with data from the test environments.
We train three versions of the MLP-BC policy, and evaluate them in the environments described in Section~\ref{sec:test_environments}. The policies are denoted as

\vspace{0.2cm}
\noindent\textbf{Zero-shot:}
Trained with the full train split of the cleaned FrodoBots8K dataset, comprising \SI{3600}{\hour} from across 153 locations. Deployed zero-shot to unseen environments.

\vspace{0.05cm}
\noindent\textbf{Scale + in-domain:}
A policy trained with the full train split \emph{plus} \SI{25}{\hour} of demonstrations from each test location.
For all environments except Selebi Phikwe, the in-domain data includes demonstrations of the route used in the experiments.

\vspace{0.05cm}
\noindent\textbf{In-domain only:}
A small-data baseline - separate policy for each test location, trained exclusively on \SI{25}{\hour} of demonstrations from the corresponding environment.

\vspace{0.2cm}
\noindent
\textbf{Results.}
Figure~\ref{fig:generalization} shows the navigation success rates, and Figure~\ref{fig:generalization_path_trace} demonstrates the policy paths over one repetition in the Wuhan environment.
The environment-specific policies overfit to the training routes. They tend to ignore the goal vector, and instead try to repeat the route seen during training.
This is beneficial on routes like Kisumu, where the robot never visits the same place twice. In Wuhan, where the robot visits the environment's central location multiple times with different goals, the policy often starts driving down a wrong part of the route.
For a reason we could not determine, the policy specific to Selebi Phikwe only drives straight, resulting in zero success.

The policies trained with data from a large number of different locations perform better.
The zero-shot policy is capable of complex behaviors like avoiding obstacles, stopping for humans, and finding intersections.
Generally, success 
\ifRALfinal
\begin{figure}[!hb]
    \centering
    \includegraphics[width=0.9\linewidth]{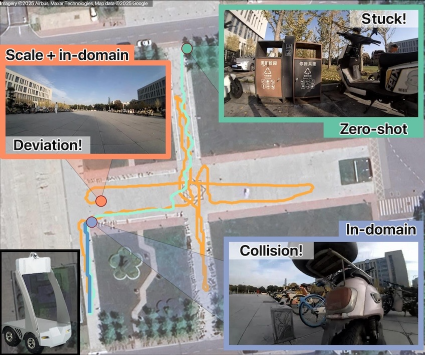}
    \caption{Policy paths until the first failure in Wuhan. See Fig.~\ref{fig:wuhan} for the mission checkpoints.}
    \label{fig:generalization_path_trace}
    \vspace{-15pt}
\end{figure}
\begin{wrapfigure}{r}{0.5\linewidth}
    \centering
    \vspace{-5pt} 
    \includegraphics[width=\linewidth, trim={0 2,5cm 0 3,5cm}, clip]{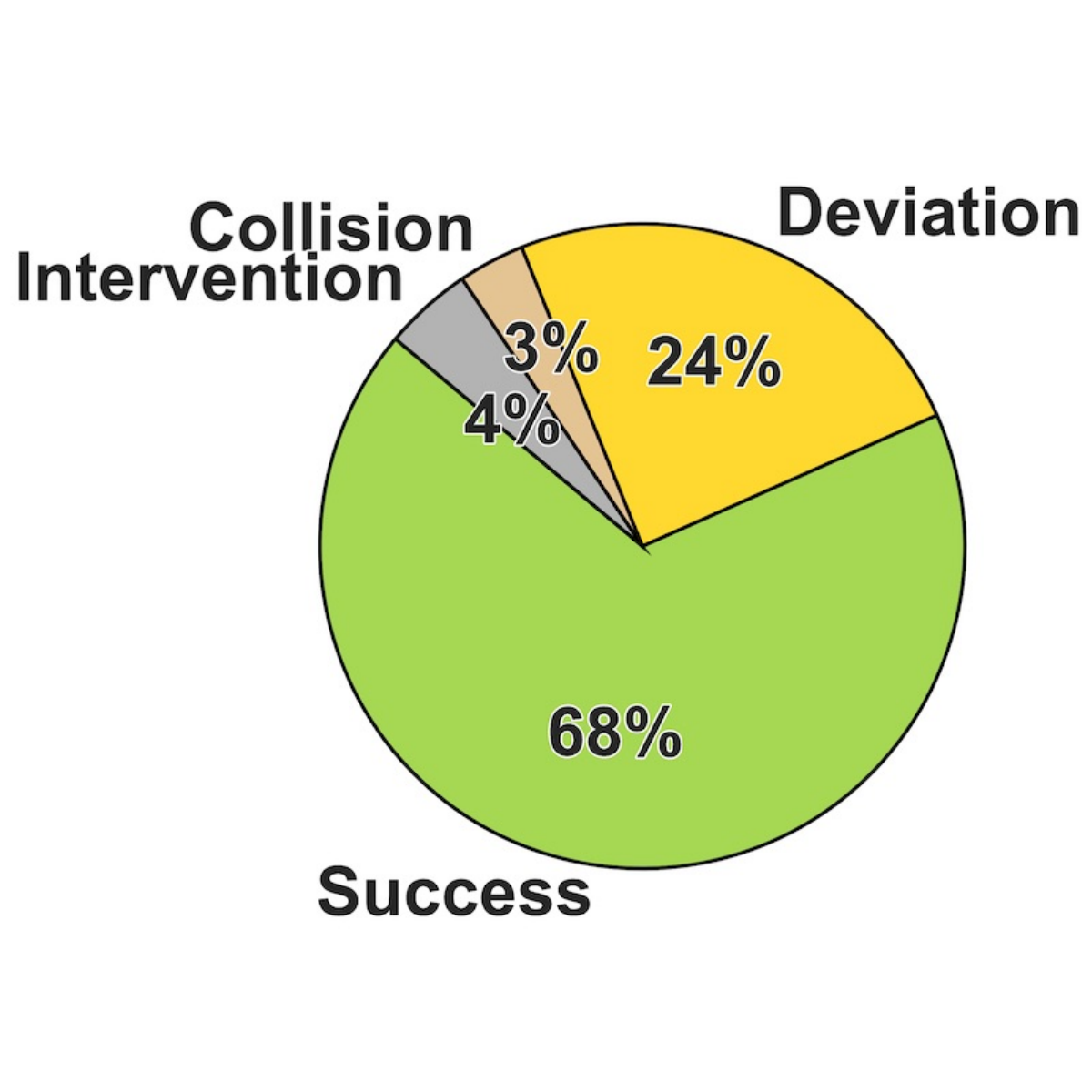}
    \caption{Failure reasons for the zero-shot policy.}
    \label{fig:failure_reasons}
    \vspace{-15pt}
\end{wrapfigure}
\fi
rates of the zero-shot and in-domain versions of the large-scale policy are very close.
As shown by Figure~\ref{fig:failure_reasons}, 
\ifpreprint
\begin{figure}[!hb]
    \centering
    \includegraphics[width=0.9\linewidth]{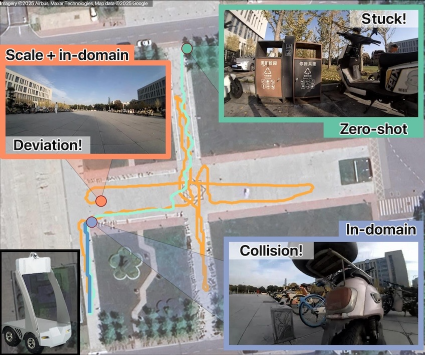}
    \caption{Policy paths until the first failure in Wuhan. See Fig.~\ref{fig:wuhan} for the mission checkpoints.}
    \label{fig:generalization_path_trace}
    \vspace{-15pt}
\end{figure}
\begin{wrapfigure}{r}{0.5\linewidth}
    \centering
    \vspace{-5pt} 
    \includegraphics[width=\linewidth, trim={0 2,5cm 0 3,5cm}, clip]{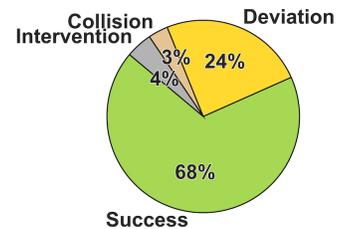}
    \caption{Failure reasons for the zero-shot policy.}
    \label{fig:failure_reasons}
    \vspace{-15pt}
\end{wrapfigure}
\fi
the most common failure mode for the zero-shot policy is deviating from the route~\eg~because of taking a wrong turn at an intersection or failing to perform a \SI{180}{\degree} turn after goal change.
We hypothesize two potential causes for this behavior, the noisy robot orientation in the training data, and the absence of longer-term planning in the point goal formulation of the navigation task.
The metrics in Table~\ref{tab:gen_additional} provide additional insight into policy performance. Incorporating large-scale data brings benefits across all metrics except progress speed, which mostly depends on the magnitude of linear velocities that a policy commands.

Overall, these results show that large-scale training data enables robust, non-trivial navigation across diverse environments. Scale gives rise to emergent behaviors, such as obstacle avoidance and stopping for pedestrians, that are absent in small-scale in-domain policies.
Moreover, the small performance gap between the large-scale zero-shot and in-domain policies suggests that sufficiently diverse training data can largely remove the need for collecting demonstrations in the deployment environment.
Next, we turn to study how the individual factors of large-scale data contribute to zero-shot navigation performance.

\begin{table}[]
    \centering
    \begin{tabular}{l c c c}
        \toprule
        Policy & NIR~$(\downarrow)$ & NPS~$(\downarrow)$ & Dist.~$(\uparrow)$\\
        \midrule
        Zero-shot & 0.86 & \SI{226}{\second} & \SI{64}{\meter} \\
        Scale + in-domain & 0.67 & \SI{220}{\second} & \SI{123}{\meter} \\
        In-domain only & 1.86 & \SI{210}{\second} & \SI{39}{\meter} \\
        \bottomrule
    \end{tabular}
    \caption{Additional metrics for the first experiment. Averages over all test environments.}
    \label{tab:gen_additional}
\end{table}

\subsection{Data diversity beats absolute scale.}
\label{sec:exp_scaling}
To examine \textbf{Q2.}, we designed an experiment to disentangle how training data \emph{size} and \emph{geographic diversity} independently contribute to zero-shot navigation performance in unseen environments.
We use the number of distinct locations in the training set as a proxy for data diversity, as different locations naturally induce variation in visual appearance, scene layout, navigation maneuvers, and route structure.
We train different versions of the MLP-BC policy with subsets of the full \SI{3600}{\hour} dataset, varying \textbf{1)} the number of geographic locations from which training data is sampled, and \textbf{2)} the amount of data sampled from each location.
In both cases, we ensure the subsets of increasing size are supersets of the smaller ones.
The feasible combinations of location count and per-location data are restricted by the composition of the full dataset.
In total, we train 21 different policies, and evaluate them on each test route 3 times.

\begin{figure}[t]
    \centering
    \includegraphics[width=1.0\linewidth]{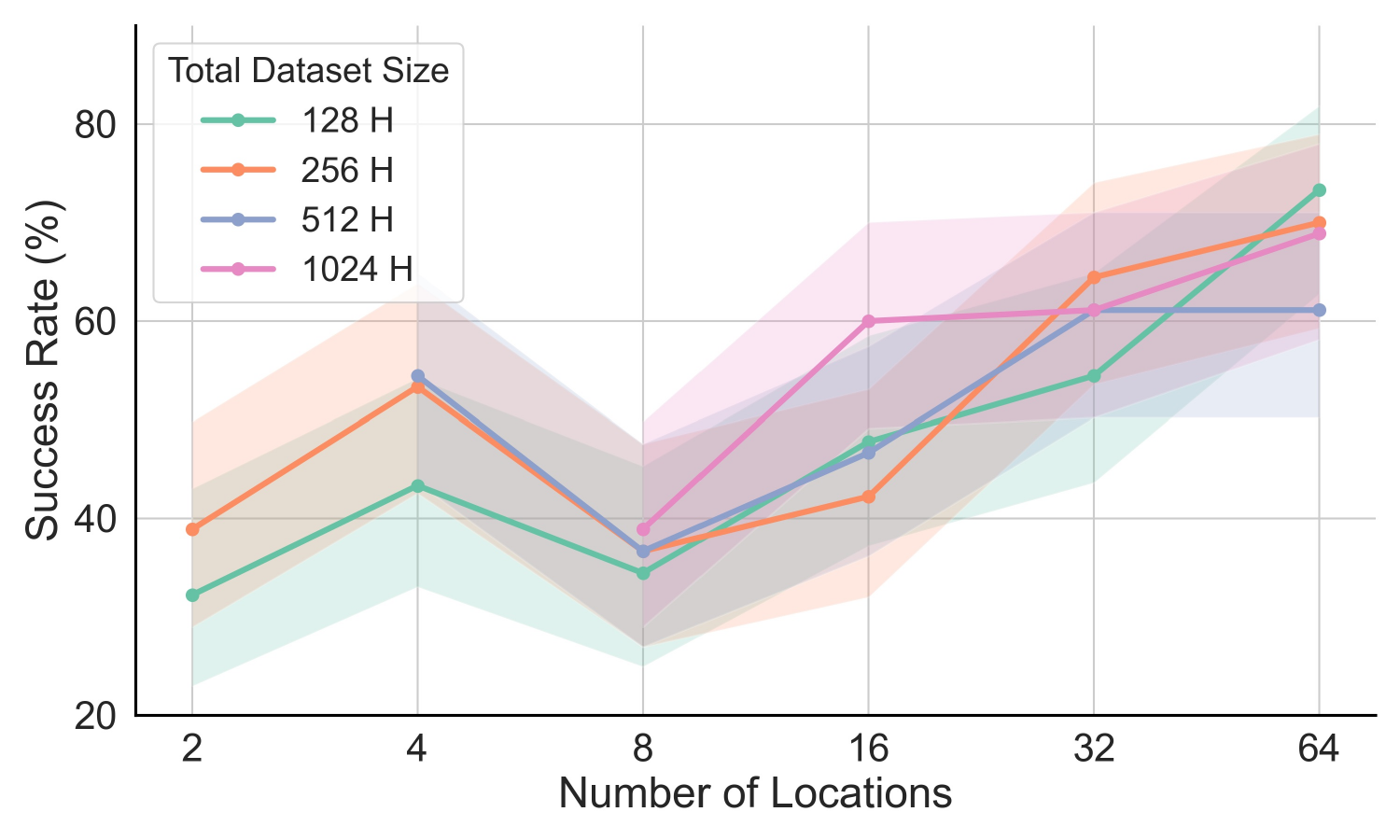}
    \caption{\emph{Success rate}~(\(\uparrow\)) as function of number of train locations, for \textbf{fixed amounts of total training data}.}
    \label{fig:fixed_totH}
    \vspace{-0.3cm}
\end{figure}

\begin{figure}[t]
    \centering
    \includegraphics[width=1.0\linewidth]{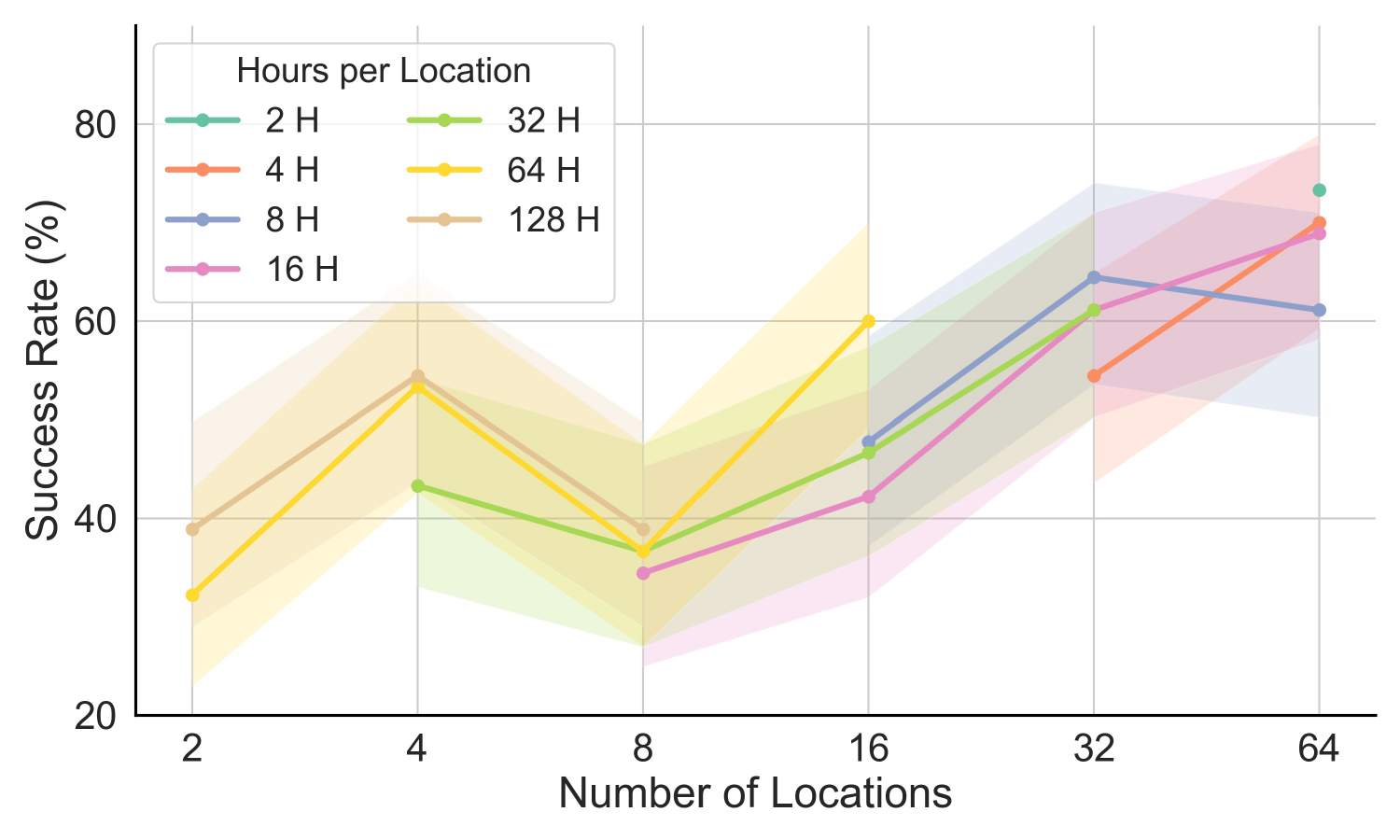}
    \caption{\emph{Success rate}~(\(\uparrow\)) as function of number of train locations, for \textbf{fixed amounts of training data per location}.}
    \label{fig:fixed_maxH}
    \vspace{-0.3cm}
\end{figure}

\begin{figure}[t]
    \centering
    \includegraphics[width=1.0\linewidth]{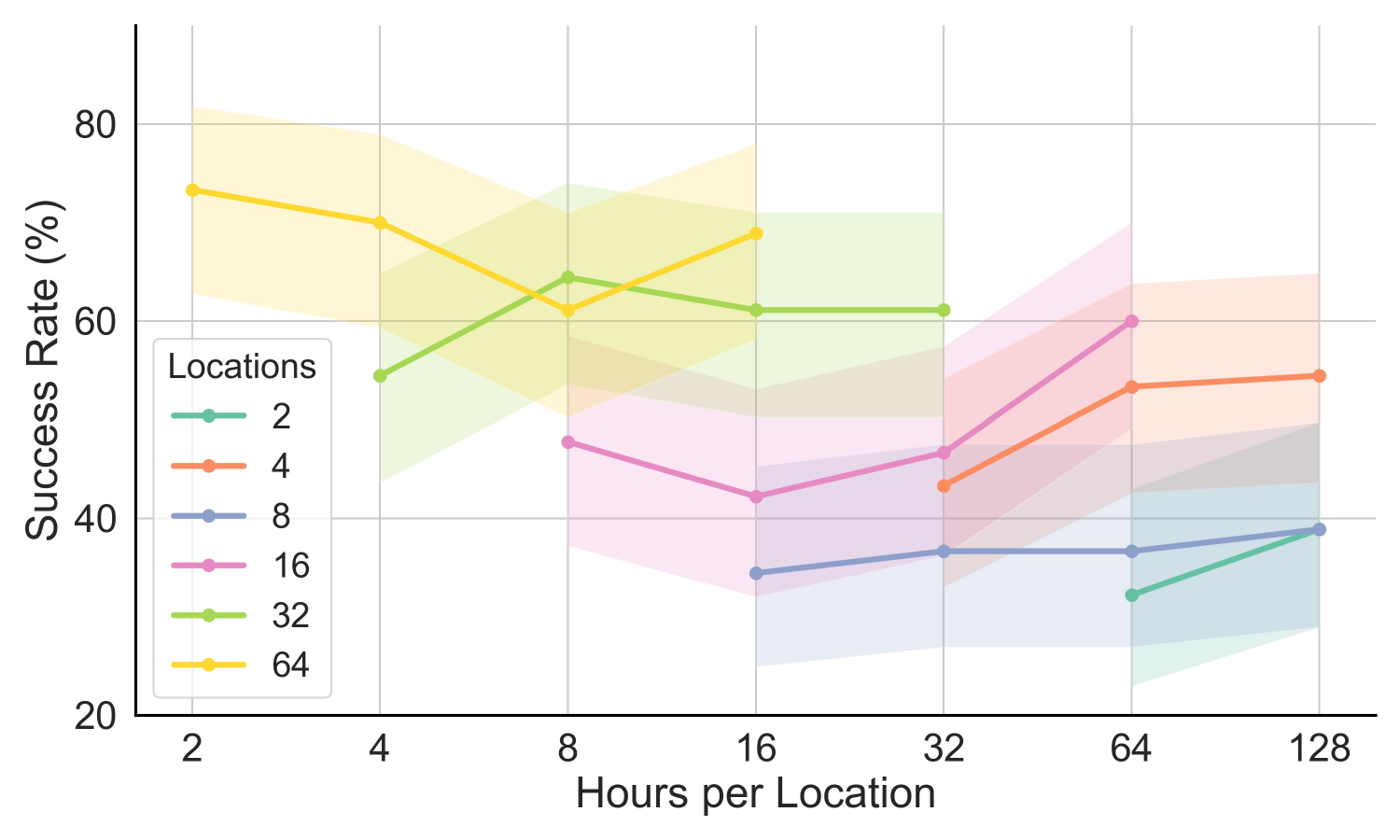}
    \caption{\emph{Success rate}~(\(\uparrow\)) as function of training data per location, for \textbf{fixed numbers of locations}.}
    \label{fig:fixed_nLoc}
    \vspace{-0.3cm}
\end{figure}

\vspace{0.2cm}
\noindent
\textbf{Results.}
In general, the policy success rates increase with additional training locations. This holds both in the case where the total dataset size is held constant (Fig.~\ref{fig:fixed_totH}), and the case where the total dataset size increases with new locations (Fig.~\ref{fig:fixed_maxH}). As seen from Fig.~\ref{fig:fixed_nLoc}, increasing the quantity of training data from any fixed set of environments has negligible effect on policy performance.

We observe an anomaly in the performance of policies trained with data from 8 locations, suggesting that the data quality across the locations is not equal. This specific mix of locations causes the policies to get stuck in infinite spinning loops in the Kisumu environment, leading to a dip in the overall success rates. The effect, however, is remedied when the number of training locations is further increased.

The results demonstrate that for navigation performance in unknown environments, it is more beneficial to increase training data size by gathering data from as many different environments as possible, rather than collect large amounts of data from few environments. Fig.~\ref{fig:fixed_nLoc} suggests that the performance increase is already saturated at \SI{2}{\hour} per location, below the range of quantities we considered in the experiment. At 64 environments, the trend of performance increase from new locations does not show signs of saturation.

\begin{figure}[t]
    \centering
    \includegraphics[width=1.0\linewidth]{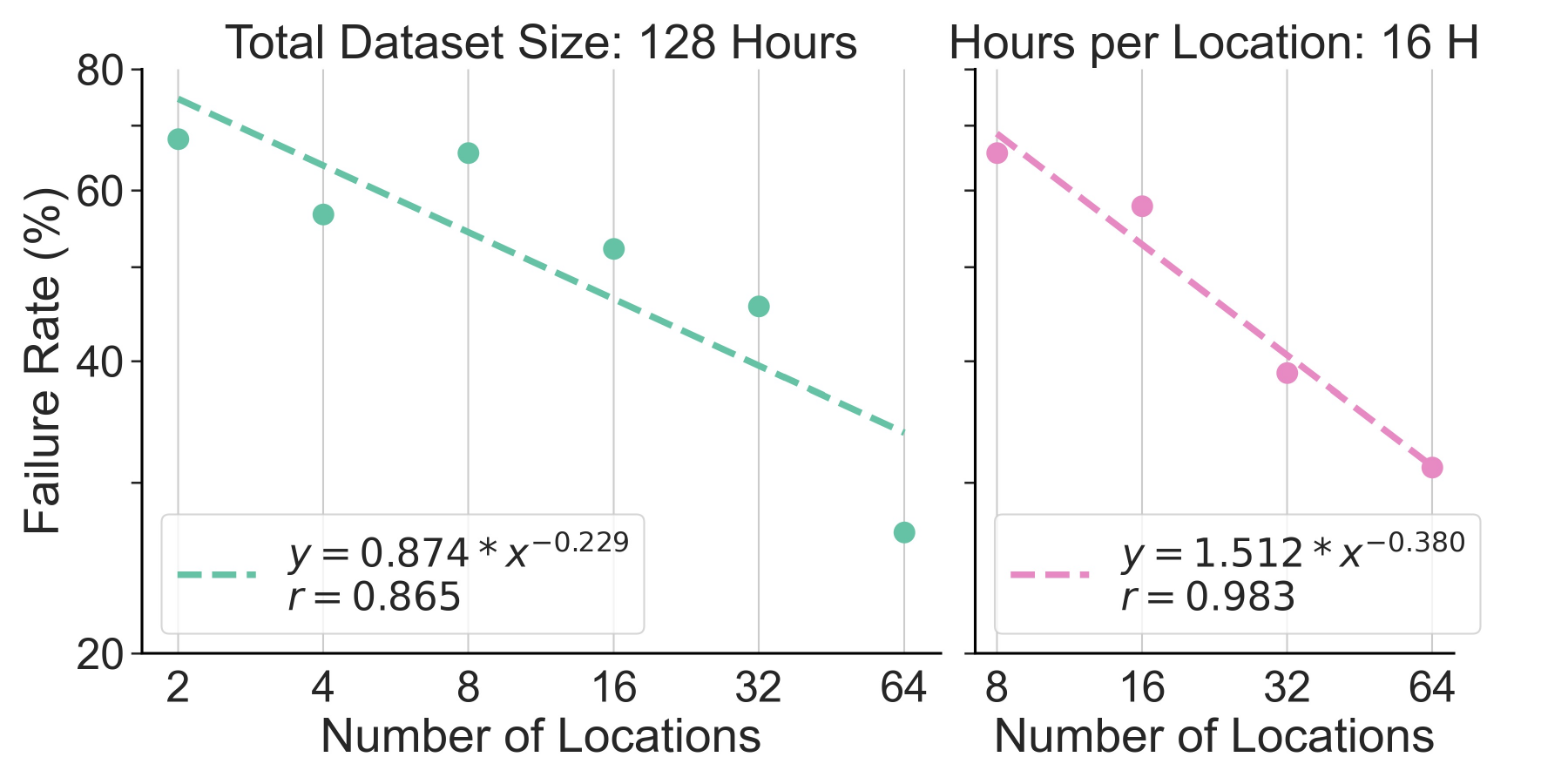}
    \caption{Power law fits and correlation coefficients \(r\) for \emph{Total Dataset Size = \SI{128}{\hour}} (Fig.~\ref{fig:fixed_totH}) and
    \emph{Hours per Location = \SI{16}{\hour}} (Fig.~\ref{fig:fixed_maxH}) on a log-log scale.}
    \label{fig:scaling_laws}
    \vspace{-0.3cm}
\end{figure}

To further analyze the connection between data diversity and navigation performance, we examine if they exhibit the same power-law relationship \(Y = \beta \cdot X^{\alpha}\) observed in data scaling studies in other fields~\cite{kaplan_scaling_2020, lin_data_2025}.
We fit a linear model on the log-transformed number of locations and the navigation \(Failure~rate = (1-Succes~rate)\). The results shown in Fig.~\ref{fig:scaling_laws} do demonstrate the existence of a power-law relationship, indicated by high values of Pearson's \(r\). While variation in the data quality between different environments causes the fit not to be perfect, the overall trend of navigation failures decreasing proportional to increase in number of train locations is clear.
Substituting the coefficients from the left panel in Fig.~\ref{fig:scaling_laws}, doubling the number of locations decreases failures by \(1-\frac{Y(2X)}{Y(X)}=1-2^{-0.229}\approx15\%\).

\subsection{With noisy data, simple models can be the most effective.}
\label{sec:exp_ablation}

In order to answer \textbf{Q3.}, we compared different policy architectures trained with regression and generative objectives.
We emphasize that we are \emph{not} proposing a new navigation method - we seek to determine if utilization of crowd-sourced demonstrations or remote deployment on low-cost robots impact the policy choice for the main experiments in Sections~\ref{sec:exp_generalization} and~\ref{sec:exp_scaling}.
Some are navigation-specific architectures from earlier work~\cite{shahViNTLargeScaleMultiTask2023, hirose_learning_2025, sridhar_nomad_2024}, and others are general robot policies~\cite{chi_diffusion_2023, black__2025}. We compare both single-observation models (\(P=1\)) and ones that utilize sequences of historical observations (\(P>1\)). We train all policies with a total of \SI{1024}{\hour} of data from 32 different locations.
All the policies utilize the Theia-Base~\cite{shang_theia_2024} vision encoder unless stated otherwise.

\vspace{0.2cm}
\noindent\textbf{MLP-BC:}
The latest image and goal embeddings are simply concatenated and passed to an MLP head to produce a chunk of \(H\) future actions \(\mathbf{a}_t\). Trained with a regression objective.

\vspace{0.1cm}
\noindent\textbf{Diffusion Policy (DP)~\cite{chi_diffusion_2023}:}
The concatenated observation embedding conditions a diffusion denoising U-Net that maps Gaussian noise into an action chunk \(\mathbf{a}_t\). Deployed with temporal ensembling~\cite{zhao_learning_2023} to reduce mode-switching.

\vspace{0.1cm}
\noindent\textbf{Flow-matching Policy (FM):}
Same architecture as DP, but trained with a flow-matching objective~\cite{black__2025}. Instead of temporal ensembling, the policy is deployed with real-time chunking~\cite{black_real-time_2025} using pseudo-inverse guidance~\cite{song_pseudoinverse-guided_2022}.

\vspace{0.1cm}
\noindent\textbf{ViNT\(^*\)~\cite{shahViNTLargeScaleMultiTask2023} / LogoNav~\cite{hirose_learning_2025}:}
The latest goal embedding and 6 latest image embeddings are processed by a non-causal transformer, and an MLP head maps an output token into an action chunk. Trained with a regression objective.

\vspace{0.1cm}
\noindent\textbf{NoMAD\(^*\)~\cite{sridhar_nomad_2024}:}
Same architecture as ViNT\(^*\), but instead of an MLP head trained for regression, a denoising U-Net~\cite{chi_diffusion_2023} conditioned with the observation sequence produces the action chunk.

\vspace{0.2cm}
\noindent
\textbf{Results.}
\begin{figure}
    \centering
    \includegraphics[width=1.0\linewidth]{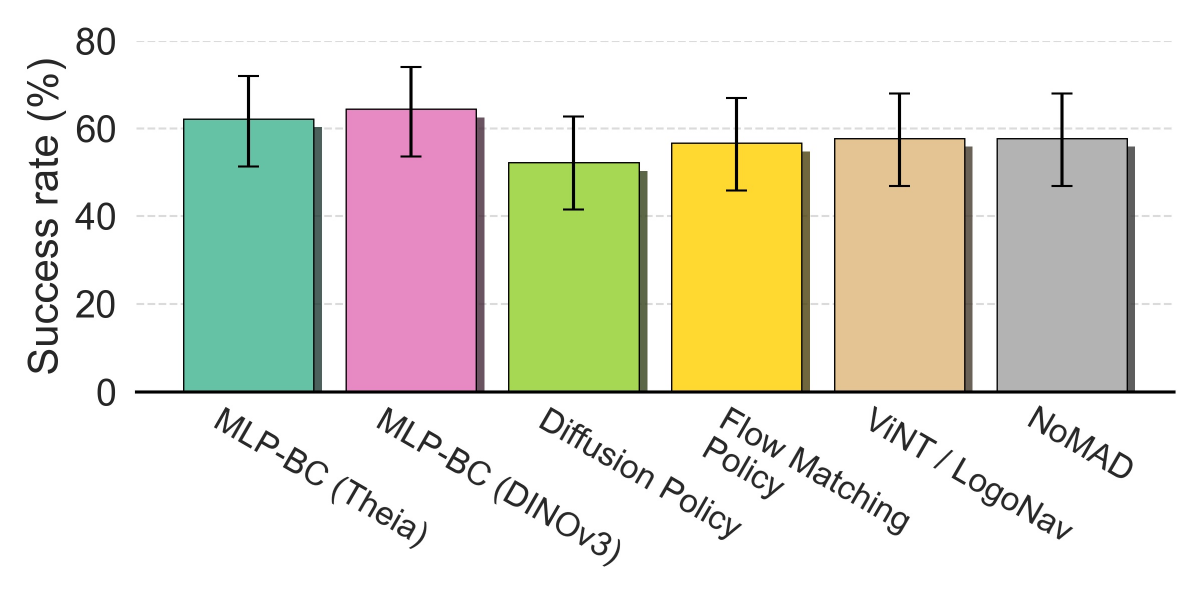}
    \caption{Comparison of policy architecture \emph{success rates}~(\(\uparrow\)), averaged over the segments for 3 repetitions in all test environments.}
    \label{fig:model_ablation}
    \vspace{-0.5cm}
\end{figure}
Figure~\ref{fig:model_ablation} shows the comparison results.
Overall, even though different methods perform worse in some test environments and better in others, there is surprisingly little difference in the average success rates.
Interestingly, the MLP-BC policies perform the best. As also observed in~\cite{kim_fine-tuning_2025}, we hypothesize that the noisy sensors and crowd-sourced demonstrations cause the generative policies (DP, FM, NoMAD) capable of modeling multi-modal action distributions to also fit the noise in the data, decreasing performance during deployment. The regression objective causes the models fit a single mode of actions, reducing the effect of noisy data.
ViNT\(^*\), which is very similar to the MLP-BC policies but utilizes a sequence of latest observations, performs worse despite being trained with a regression objective. From qualitative observations it seems like the policy learns to exploit the temporal correlation between subsequent observations too much, decreasing the policy's reactivity to~\eg~obstacles.
In addition, the variable FPS and control loop latency caused by running the policies remotely can be an issue for the sequence models.
Finally, we observe very little difference between the MLP-BC policies utilizing Theia~\cite{shang_theia_2024} or DINOv3~\cite{simeoni_dinov3_2025}, suggesting that the visual encoder is not the primary performance bottleneck.

\section{DISCUSSION \& FUTURE WORK}
We show that training end-to-end visual navigation policies on large-scale data enables generalization to unseen environments across different geographic regions. These policies significantly outperform environment-specific, in-domain policies, even when evaluated in a zero-shot manner. Our results highlight geographic diversity as a main driver of generalization. While performance gains from adding more data from a fixed set of locations saturate quickly, increasing the number of distinct training locations reduces navigation errors following a power-law relationship.

These results suggest several promising directions for future research. Extending the analysis to other domains, such as off-road navigation or aerial robotics, would help assess the generality of our findings. We also observed variation in the utility of data across training environments, as reflected in downstream policy performance. A more fine-grained understanding of the underlying causes of this variation is an important direction for future work - what kind of demonstrations are the best for learning navigation? Overall, we hope these findings motivate continued investigation into dataset composition and data scaling strategies for navigation and robot learning.




\section*{Limitations}

Our experiments were conducted in only 4 different countries and environments. Verifying the findings on a set of test environments an order of magnitude larger would help confirm their validity.
The study was performed in the context of PointGoal navigation, without the use of prior maps or deployment time 'memory'. This limits the difficulty level of the navigation tasks we considered - kilometer-scale segments that require planning, and exploration with dead-ends, for example, are challenging for such 'reactive' policies.

\section*{ACKNOWLEDGMENT}

We wish to thank the FrodoBotsAI team for operations support and generous access to their robot fleet.
We acknowledge CSC, IT Center for Science, Finland, for awarding this project access to the LUMI supercomputer, owned by the EuroHPC Joint Undertaking, hosted by CSC (Finland) and the LUMI consortium.


\bibliographystyle{IEEEtran_nourl}
\bibliography{references}

@inproceedings{shahViNTLargeScaleMultiTask2023,
    title = {{ViNT}: {A} {Large}-{Scale}, {Multi}-{Task} {Visual} {Navigation} {Backbone} with {Cross}-{Robot} {Generalization}},
    shorttitle = {{ViNT}},
    url = {https://openreview.net/forum?id=-K7-1WvKO3F},
    language = {en},
    urldate = {2023-09-08},
    booktitle = {7th Annual Conference on {Robot} {Learning}},
    author = {Shah, Dhruv and Sridhar, Ajay and Dashora, Nitish and Stachowicz, Kyle and Black, Kevin and Hirose, Noriaki and Levine, Sergey},
    month = aug,
    year = {2023},
}

@article{lin_data_2025,
    title = {Data {Scaling} {Laws} in {Imitation} {Learning} for {Robotic} {Manipulation}},
    url = {https://proceedings.iclr.cc/paper_files/paper/2025/hash/88b7b2c896506daabc8d3fd587055167-Abstract-Conference.html},
    language = {en},
    urldate = {2025-10-03},
    journal = {International Conference on Representation Learning},
    author = {Lin, Fanqi and Hu, Yingdong and Sheng, Pingyue and Wen, Chuan and You, Jiacheng and Gao, Yang},
    month = may,
    year = {2025},
    pages = {54877--54910},
}

@inproceedings{black_pi05_2025,
    title={\(\pi_{0.5}\): a Vision-Language-Action Model with Open-World Generalization},
    shorttitle = {pi0.5},
    url = {https://openreview.net/forum?id=vlhoswksBO#discussion},
    language = {en},
    urldate = {2025-10-09},
    author = {Black, Kevin and Brown, Noah and Darpinian, James and Dhabalia, Karan and Driess, Danny and Esmail, Adnan and Equi, Michael Robert and Finn, Chelsea and Fusai, Niccolo and Galliker, Manuel Y. and Ghosh, Dibya and Groom, Lachy and Hausman, Karol and Ichter, Brian and Jakubczak, Szymon and Jones, Tim and Ke, Liyiming and LeBlanc, Devin and Levine, Sergey and Li-Bell, Adrian and Mothukuri, Mohith and Nair, Suraj and Pertsch, Karl and Ren, Allen Z. and Shi, Lucy Xiaoyang and Smith, Laura and Springenberg, Jost Tobias and Stachowicz, Kyle and Tanner, James and Vuong, Quan and Walke, Homer and Walling, Anna and Wang, Haohuan and Yu, Lili and Zhilinsky, Ury},
    month = sep,
    year = {2025},
    booktitle={9th Annual Conference on {Robot} Learning},
}

@misc{zheng_preliminary_2024,
    title = {Preliminary {Investigation} into {Data} {Scaling} {Laws} for {Imitation} {Learning}-{Based} {End}-to-{End} {Autonomous} {Driving}},
    url = {http://arxiv.org/abs/2412.02689},
    doi = {10.48550/arXiv.2412.02689},
    language = {en},
    urldate = {2025-10-09},
    publisher = {arXiv},
    author = {Zheng, Yupeng and Xia, Zhongpu and Zhang, Qichao and Zhang, Teng and Lu, Ben and Huo, Xiaochuang and Han, Chao and Li, Yixian and Yu, Mengjie and Jin, Bu and Yang, Pengxuan and Zheng, Yuhang and Yuan, Haifeng and Jiang, Ke and Jia, Peng and Lang, Xianpeng and Zhao, Dongbin},
    month = dec,
    year = {2024},
    note = {arXiv:2412.02689 [cs]}
}

@inproceedings{liu_citywalker_2025,
    title = {{CityWalker}: {Learning} {Embodied} {Urban} {Navigation} from {Web}-{Scale} {Videos}},
    shorttitle = {{CityWalker}},
    url = {https://openaccess.thecvf.com/content/CVPR2025/html/Liu_CityWalker_Learning_Embodied_Urban_Navigation_from_Web-Scale_Videos_CVPR_2025_paper.html},
    booktitle = {IEEE/CVF Conference on Computer Vision and Pattern Recognition},
    language = {en},
    urldate = {2025-10-09},
    author = {Liu, Xinhao and Li, Jintong and Jiang, Yicheng and Sujay, Niranjan and Yang, Zhicheng and Zhang, Juexiao and Abanes, John and Zhang, Jing and Feng, Chen},
    year = {2025},
    pages = {6875--6885},
}

@inproceedings{etukuru_robot_2025,
    title = {Robot {Utility} {Models}: {General} {Policies} for {Zero}-{Shot} {Deployment} in {New} {Environments}},
    shorttitle = {Robot {Utility} {Models}},
    url = {https://ieeexplore.ieee.org/abstract/document/11127857},
    doi = {10.1109/ICRA55743.2025.11127857},
    urldate = {2025-10-09},
    booktitle = {{IEEE} {International} {Conference} on {Robotics} and {Automation}},
    author = {Etukuru, Haritheja and Naka, Norihito and Hu, Zijin and Lee, Seungjae and Mehu, Julian and Edsinger, Aaron and Paxton, Chris and Chintala, Soumith and Pinto, Lerrel and Mahi Shafiullah, Nur Muhammad},
    month = may,
    year = {2025},
    pages = {8275--8283},
}

@misc{sartor_neural_2025,
    title = {Neural {Scaling} {Laws} in {Robotics}},
    url = {http://arxiv.org/abs/2405.14005},
    doi = {10.48550/arXiv.2405.14005},
    urldate = {2025-10-15},
    publisher = {arXiv},
    author = {Sartor, Sebastian and Thompson, Neil},
    month = jan,
    year = {2025},
    note = {arXiv:2405.14005 [cs]},
}

@inproceedings{hirose_learning_2025,
    title = {Learning to {Drive} {Anywhere} with {Model}-{Based} {Reannotation}},
    url = {https://openreview.net/forum?id=9DyLaIHqrD},
    language = {en},
    booktitle = {ICRA Workshops},
    urldate = {2025-10-22},
    author = {Hirose, Noriaki and Ignatova, Lydia and Stachowicz, Kyle and Glossop, Catherine and Levine, Sergey and Shah, Dhruv},
    month = may,
    year = {2025},
}

@inproceedings{shah_viking_2022,
    title = {{ViKiNG}: {Vision}-{Based} {Kilometer}-{Scale} {Navigation} with {Geographic} {Hints}},
    isbn = {978-0-9923747-8-5},
    booktitle = {Robotics: Science and Systems XVIII},
    shorttitle = {{ViKiNG}},
    url = {https://www.roboticsproceedings.org/rss18/p019.html},
    urldate = {2025-10-22},
    author = {Shah, Dhruv and Levine, Sergey},
    month = jun,
    year = {2022},
}

@inproceedings{suomela_placenav_2024,
    title = {{PlaceNav}: {Topological} {Navigation} through {Place} {Recognition}},
    shorttitle = {{PlaceNav}},
    url = {https://ieeexplore.ieee.org/document/10610575},
    doi = {10.1109/ICRA57147.2024.10610575},
    urldate = {2024-08-22},
    booktitle = {{IEEE} {International} {Conference} on {Robotics} and {Automation}},
    author = {Suomela, Lauri and Kalliola, Jussi and Edelman, Harry and Kämäräinen, Joni-Kristian},
    month = may,
    year = {2024},
    pages = {5205--5213},
}

@misc{hirose_omnivla_2025,
    title = {{OmniVLA}: {An} {Omni}-{Modal} {Vision}-{Language}-{Action} {Model} for {Robot} {Navigation}},
    shorttitle = {{OmniVLA}},
    url = {http://arxiv.org/abs/2509.19480},
    doi = {10.48550/arXiv.2509.19480},
    urldate = {2025-10-22},
    publisher = {arXiv},
    author = {Hirose, Noriaki and Glossop, Catherine and Shah, Dhruv and Levine, Sergey},
    month = sep,
    year = {2025},
    note = {arXiv:2509.19480 [cs]},
}

@article{shen_effonav_2025,
    title = {{EffoNAV}: {An} {Effective} {Foundation}-{Model}-{Based} {Visual} {Navigation} {Approach} in {Challenging} {Environment}},
    volume = {10},
    issn = {2377-3766},
    shorttitle = {{EffoNAV}},
    url = {https://ieeexplore.ieee.org/document/11012733},
    doi = {10.1109/LRA.2025.3572815},
    number = {7},
    urldate = {2025-10-22},
    journal = {IEEE Robotics and Automation Letters},
    author = {Shen, Wangtian and Gu, Pengfei and Qin, Haijian and Meng, Ziyang},
    month = jul,
    year = {2025},
    pages = {6824--6831},
}

@inproceedings{ehsani_spoc_2024,
    title = {{SPOC}: {Imitating} {Shortest} {Paths} in {Simulation} {Enables} {Effective} {Navigation} and {Manipulation} in the {Real} {World}},
    shorttitle = {{SPOC}},
    url = {https://openaccess.thecvf.com/content/CVPR2024/html/Ehsani_SPOC_Imitating_Shortest_Paths_in_Simulation_Enables_Effective_Navigation_and_CVPR_2024_paper.html},
    language = {en},
    urldate = {2024-08-30},
    booktitle = {{IEEE}/{CVF} {Conference} on {Computer} {Vision} and {Pattern} {Recognition}},
    author = {Ehsani, Kiana and Gupta, Tanmay and Hendrix, Rose and Salvador, Jordi and Weihs, Luca and Zeng, Kuo-Hao and Singh, Kunal Pratap and Kim, Yejin and Han, Winson and Herrasti, Alvaro and Krishna, Ranjay and Schwenk, Dustin and VanderBilt, Eli and Kembhavi, Aniruddha},
    year = {2024},
    pages = {16238--16250},
}

@inproceedings{shahGNMGeneralNavigation2023,
    title = {{GNM}: {A} {General} {Navigation} {Model} to {Drive} {Any} {Robot}},
    shorttitle = {{GNM}},
    doi = {10.1109/ICRA48891.2023.10161227},
    booktitle = {{IEEE} {International} {Conference} on {Robotics} and {Automation}},
    author = {Shah, Dhruv and Sridhar, Ajay and Bhorkar, Arjun and Hirose, Noriaki and Levine, Sergey},
    month = may,
    year = {2023},
    pages = {7226--7233},
}

@inproceedings{hirose_lelan_2024,
    title = {{LeLaN}: {Learning} {A} {Language}-{Conditioned} {Navigation} {Policy} from {In}-the-{Wild} {Video}},
    shorttitle = {{LeLaN}},
    url = {https://openreview.net/forum?id=zIWu9Kmlqk},
    booktitle = {8th Annual Conference on Robot Learning},
    language = {en},
    urldate = {2025-10-22},
    author = {Hirose, Noriaki and Glossop, Catherine and Sridhar, Ajay and Mees, Oier and Levine, Sergey},
    month = sep,
    year = {2024},
}

@misc{noauthor_lerobot_2025,
    author = {{Yaak \& LeRobot team}},
    title = {{LeRobot} goes to driving school: {World}’s largest open-source self-driving dataset},
    note = {\url{https://www.huggingface.com/blog/lerobot-goes-to-driving-school}},
    shorttitle = {{LeRobot} goes to driving school},
    url = {https://huggingface.co/blog/lerobot-goes-to-driving-school},
    urldate = {2025-10-23},
    year = {2025},
}

@inproceedings{muller_openbot-fleet_2024,
    title = {{OpenBot}-{Fleet}: {A} {System} for {Collective} {Learning} with {Real} {Robots}},
    shorttitle = {{OpenBot}-{Fleet}},
    url = {https://ieeexplore.ieee.org/document/10610960},
    doi = {10.1109/ICRA57147.2024.10610960},
    urldate = {2024-09-18},
    booktitle = {{IEEE} {International} {Conference} on {Robotics} and {Automation}},
    author = {Müller, Matthias and Brahmbhatt, Samarth and Deka, Ankur and Leboutet, Quentin and Hafner, David and Koltun, Vladlen},
    month = may,
    year = {2024},
    pages = {4758--4765},
}

@inproceedings{xu_end--end_2017,
    title = {End-to-{End} {Learning} of {Driving} {Models} from {Large}-{Scale} {Video} {Datasets}},
    isbn = {978-1-5386-0457-1},
    url = {https://www.computer.org/csdl/proceedings-article/cvpr/2017/0457d530/12OmNAZx8L2},
    doi = {10.1109/CVPR.2017.376},
    booktitle = {IEEE/CVF Conference on Computer Vision and Pattern Recognition},
    language = {English},
    urldate = {2025-10-23},
    author = {Xu, Huazhe and Gao, Yang and Yu, Fisher and Darrell, Trevor},
    month = jul,
    year = {2017},
    pages = {3530--3538},
}

@misc{frodobots_frodobots-2k_nodate,
    title = {{FrodoBots}-{2K} Dataset},
    note = {\url{https://huggingface.co/datasets/frodobots/FrodoBots-2K}},
    urldate = {2025-10-23},
    year = {2025},
}

@inproceedings{sridhar_nomad_2024,
    title = {{NoMaD}: {Goal} {Masked} {Diffusion} {Policies} for {Navigation} and {Exploration}},
    shorttitle = {{NoMaD}},
    url = {https://ieeexplore.ieee.org/abstract/document/10610665},
    doi = {10.1109/ICRA57147.2024.10610665},
    urldate = {2024-08-22},
    booktitle = {{IEEE} {International} {Conference} on {Robotics} and {Automation}},
    author = {Sridhar, Ajay and Shah, Dhruv and Glossop, Catherine and Levine, Sergey},
    month = may,
    year = {2024},
    pages = {63--70},
}

@inproceedings{chi_diffusion_2023,
    title = {Diffusion {Policy}: {Visuomotor} {Policy} {Learning} via {Action} {Diffusion}},
    isbn = {978-0-9923747-9-2},
    shorttitle = {Diffusion {Policy}},
    booktitle = {Robotics: Science and Systems XIX},
    url = {https://roboticsproceedings.org/rss19/p026.html},
    urldate = {2025-10-24},
    author = {Chi, Cheng and Feng, Siyuan and Du, Yilun and Xu, Zhenjia and Cousineau, Eric and Burchfiel, Benjamin CM and Song, Shuran},
    month = jul,
    year = {2023},
}

@article{anderson_evaluation_2018,
    title = {On {Evaluation} of {Embodied} {Navigation} {Agents}},
    url = {http://arxiv.org/abs/1807.06757},
    urldate = {2021-02-18},
    journal = {arXiv:1807.06757 [cs]},
    author = {Anderson, Peter and Chang, Angel and Chaplot, Devendra Singh and Dosovitskiy, Alexey and Gupta, Saurabh and Koltun, Vladlen and Kosecka, Jana and Malik, Jitendra and Mottaghi, Roozbeh and Savva, Manolis and Zamir, Amir R.},
    month = jul,
    year = {2018},
}

@inproceedings{wijmansDDPPOLearningNearPerfect2020,
    title = {{DD}-{PPO}: {Learning} {Near}-{Perfect} {PointGoal} {Navigators} from 2.5 {Billion} {Frames}},
    shorttitle = {{DD}-{PPO}},
    url = {https://openreview.net/forum?id=H1gX8C4YPr},
    urldate = {2025-01-10},
    booktitle = {International {Conference} on {Learning} {Representations}},
    author = {Wijmans, Erik and Kadian, Abhishek and Morcos, Ari and Lee, Stefan and Essa, Irfan and Parikh, Devi and Savva, Manolis and Batra, Dhruv},
    year = {2020},
}

@inproceedings{xie_vid2sim_2025,
    title = {{Vid2Sim}: {Realistic} and {Interactive} {Simulation} from {Video} for {Urban} {Navigation}},
    shorttitle = {{Vid2Sim}},
    booktitle = {IEEE/CVF Conference on Computer Vision and Pattern Recognition},
    url = {https://openaccess.thecvf.com/content/CVPR2025/html/Xie_Vid2Sim_Realistic_and_Interactive_Simulation_from_Video_for_Urban_Navigation_CVPR_2025_paper.html},
    language = {en},
    urldate = {2025-10-29},
    author = {Xie, Ziyang and Liu, Zhizheng and Peng, Zhenghao and Wu, Wayne and Zhou, Bolei},
    year = {2025},
    pages = {1581--1591},
}

@misc{frank_dellaert_and_gtsam_contributors_borglabgtsam_2022,
    title = {borglab/gtsam},
    note = {\url{https://github.com/borglab/gtsam}},
    publisher = {Georgia Tech Borg Lab},
    author = {{Frank Dellaert and GTSAM Contributors}},
    month = may,
    year = {2022},
}

@inproceedings{villasevil_robot_2025,
    title = {Robot {Learning} with {Super}-{Linear} {Scaling}},
    isbn = {979-8-9902848-1-4},
    url = {https://www.roboticsproceedings.org/rss21/p025.html},
    urldate = {2025-10-31},
    booktitle = {Robotics: Science and Systems XXI},
    author = {Villasevil, Marcel Torne and Jain, Arhan and Yuan, Jiayi and Macha, Vidyaaranya and Ankile, Lars Lien and Simeonov, Anthony and Agrawal, Pulkit and Gupta, Abhishek},
    month = jun,
    year = {2025},
}

@article{kahn_badgr_2021,
    title = {{BADGR}: {An} {Autonomous} {Self}-{Supervised} {Learning}-{Based} {Navigation} {System}},
    volume = {6},
    issn = {2377-3766},
    shorttitle = {{BADGR}},
    url = {https://ieeexplore.ieee.org/document/9345970},
    doi = {10.1109/LRA.2021.3057023},
    number = {2},
    urldate = {2024-08-22},
    journal = {IEEE Robotics and Automation Letters},
    author = {Kahn, Gregory and Abbeel, Pieter and Levine, Sergey},
    month = apr,
    year = {2021},
    pages = {1312--1319},
}

@inproceedings{frey_fast_2023,
    title = {Fast {Traversability} {Estimation} for {Wild} {Visual} {Navigation}},
    isbn = {978-0-9923747-9-2},
    booktitle = {Robotics: Science and Systems XIX},
    url = {https://www.roboticsproceedings.org/rss19/p054.html},
    urldate = {2025-10-28},
    author = {Frey, Jonas and Mattamala, Matias and Chebrolu, Nived and Cadena, Cesar and Fallon, Maurice and Hutter, Marco},
    month = jul,
    year = {2023},
}

@inproceedings{sivaprakasam_salon_2025,
    title = {{SALON}: {Self}-supervised {Adaptive} {Learning} for {Off}-road {Navigation}},
    shorttitle = {{SALON}},
    url = {https://ieeexplore.ieee.org/abstract/document/11128268},
    doi = {10.1109/ICRA55743.2025.11128268},
    urldate = {2025-11-03},
    booktitle = {{IEEE} {International} {Conference} on {Robotics} and {Automation}},
    author = {Sivaprakasam, Matthew and Triest, Samuel and Ho, Cherie and Aich, Shubhra and Lew, Jeric and Adu, Isaiah and Wang, Wenshan and Scherer, Sebastian},
    month = may,
    year = {2025},
    pages = {16999--17006},
}

@inproceedings{zhang_creste_2025,
    title = {{CREStE}: {Scalable} {Mapless} {Navigation} with {Internet} {Scale} {Priors} and {Counterfactual} {Guidance}},
    isbn = {979-8-9902848-1-4},
    booktitle = {Robotics: Science and Systems XXI},
    shorttitle = {{CREStE}},
    url = {https://www.roboticsproceedings.org/rss21/p136.html},
    urldate = {2025-11-03},
    author = {Zhang, Arthur and Sikchi, Harshit and Biswas, Joydeep and Zhang, Amy},
    month = jun,
    year = {2025},
}

@article{elfes_using_1989,
    title = {Using occupancy grids for mobile robot perception and navigation},
    volume = {22},
    issn = {1558-0814},
    url = {https://ieeexplore.ieee.org/document/30720},
    doi = {10.1109/2.30720},
    number = {6},
    urldate = {2025-11-03},
    journal = {Computer},
    author = {Elfes, A.},
    month = jun,
    year = {1989},
    pages = {46--57},
}

@article{fox_dynamic_1997,
    title = {The dynamic window approach to collision avoidance},
    volume = {4},
    issn = {1558-223X},
    url = {https://ieeexplore.ieee.org/document/580977},
    doi = {10.1109/100.580977},
    number = {1},
    urldate = {2025-11-03},
    journal = {IEEE Robotics \& Automation Magazine},
    author = {Fox, D. and Burgard, W. and Thrun, S.},
    month = mar,
    year = {1997},
    pages = {23--33},
}

@inproceedings{lu_layered_2014,
    title = {Layered costmaps for context-sensitive navigation},
    url = {https://ieeexplore.ieee.org/abstract/document/6942636},
    doi = {10.1109/IROS.2014.6942636},
    urldate = {2025-11-03},
    booktitle = {{IEEE}/{RSJ} {International} {Conference} on {Intelligent} {Robots} and {Systems}},
    author = {Lu, David V. and Hershberger, Dave and Smart, William D.},
    month = sep,
    year = {2014},
    pages = {709--715},
}

@misc{he_seeing_2025,
    title = {From {Seeing} to {Experiencing}: {Scaling} {Navigation} {Foundation} {Models} with {Reinforcement} {Learning}},
    shorttitle = {From {Seeing} to {Experiencing}},
    url = {http://arxiv.org/abs/2507.22028},
    doi = {10.48550/arXiv.2507.22028},
    urldate = {2025-11-03},
    publisher = {arXiv},
    author = {He, Honglin and Ma, Yukai and Wu, Wayne and Zhou, Bolei},
    month = jul,
    year = {2025},
    note = {arXiv:2507.22028 [cs]},
}

@misc{frodobots_frodobots_nodate,
    title = {{FrodoBots} {AI}},
    note = {\url{https://www.frodobots.ai/}},
    urldate = {2025-11-04},
}

@inproceedings{majumdar_where_2024,
    title = {Where are we in the search for an artificial visual cortex for embodied intelligence?},
    volume = {37},
    urldate = {2024-09-19},
    booktitle = {Advances in {Neural} {Information} {Processing} {Systems}},
    author = {Majumdar, Arjun and Yadav, Karmesh and Arnaud, Sergio and Ma, Yecheng Jason and Chen, Claire and Silwal, Sneha and Jain, Aryan and Berges, Vincent-Pierre and Wu, Tingfan and Vakil, Jay and Abbeel, Pieter and Malik, Jitendra and Batra, Dhruv and Lin, Yixin and Maksymets, Oleksandr and Rajeswaran, Aravind and Meier, Franziska},
    month = may,
    year = {2024},
    pages = {655--677},
}

@inproceedings{zhao_learning_2023,
    title = {Learning {Fine}-{Grained} {Bimanual} {Manipulation} with {Low}-{Cost} {Hardware}},
    isbn = {978-0-9923747-9-2},
    booktitle = {Robotics: Science and Systems XIX},
    url = {https://www.roboticsproceedings.org/rss19/p016.html},
    urldate = {2025-11-04},
    author = {Zhao, Tony Z. and Kumar, Vikash and Levine, Sergey and Finn, Chelsea},
    month = jul,
    year = {2023},
}

@inproceedings{song_pseudoinverse-guided_2022,
    title = {Pseudoinverse-{Guided} {Diffusion} {Models} for {Inverse} {Problems}},
    url = {https://openreview.net/forum?id=9_gsMA8MRKQ},
    language = {en},
    booktitle={International Conference on Learning Representations},
    urldate = {2025-11-04},
    author = {Song, Jiaming and Vahdat, Arash and Mardani, Morteza and Kautz, Jan},
    year = {2023},
}

@inproceedings{shang_theia_2024,
    title = {Theia: {Distilling} {Diverse} {Vision} {Foundation} {Models} for {Robot} {Learning}},
    shorttitle = {Theia},
    url = {https://openreview.net/forum?id=ylZHvlwUcI},
    language = {en},
    urldate = {2024-12-27},
    booktitle = {8th Annual Conference on {Robot} {Learning}},
    author = {Shang, Jinghuan and Schmeckpeper, Karl and May, Brandon B. and Minniti, Maria Vittoria and Kelestemur, Tarik and Watkins, David and Herlant, Laura},
    month = sep,
    year = {2024},
}

@misc{simeoni_dinov3_2025,
    title = {{DINOv3}},
    url = {http://arxiv.org/abs/2508.10104},
    doi = {10.48550/arXiv.2508.10104},
    urldate = {2025-11-05},
    publisher = {arXiv},
    author = {Siméoni, Oriane and Vo, Huy V. and Seitzer, Maximilian and Baldassarre, Federico and Oquab, Maxime and Jose, Cijo and Khalidov, Vasil and Szafraniec, Marc and Yi, Seungeun and Ramamonjisoa, Michaël and Massa, Francisco and Haziza, Daniel and Wehrstedt, Luca and Wang, Jianyuan and Darcet, Timothée and Moutakanni, Théo and Sentana, Leonel and Roberts, Claire and Vedaldi, Andrea and Tolan, Jamie and Brandt, John and Couprie, Camille and Mairal, Julien and Jégou, Hervé and Labatut, Patrick and Bojanowski, Piotr},
    month = aug,
    year = {2025},
    note = {arXiv:2508.10104 [cs]},
}

@article{newcombe_two-sided_1998,
    title = {Two-sided confidence intervals for the single proportion: comparison of seven methods},
    volume = {17},
    issn = {0277-6715},
    shorttitle = {Two-sided confidence intervals for the single proportion},
    doi = {10.1002/(sici)1097-0258(19980430)17:8<857::aid-sim777>3.0.co;2-e},
    language = {eng},
    number = {8},
    journal = {Statistics in Medicine},
    author = {Newcombe, R. G.},
    month = apr,
    year = {1998},
    pmid = {9595616},
    pages = {857--872},
}

@misc{kim_fine-tuning_2025,
    title = {Fine-{Tuning} {Vision}-{Language}-{Action} {Models}: {Optimizing} {Speed} and {Success}},
    shorttitle = {Fine-{Tuning} {Vision}-{Language}-{Action} {Models}},
    url = {http://arxiv.org/abs/2502.19645},
    doi = {10.48550/arXiv.2502.19645},
    urldate = {2025-12-01},
    publisher = {arXiv},
    author = {Kim, Moo Jin and Finn, Chelsea and Liang, Percy},
    month = apr,
    year = {2025},
    note = {arXiv:2502.19645 [cs]},
}

@misc{kaplan_scaling_2020,
    title = {Scaling {Laws} for {Neural} {Language} {Models}},
    url = {http://arxiv.org/abs/2001.08361},
    doi = {10.48550/arXiv.2001.08361},
    urldate = {2025-12-12},
    publisher = {arXiv},
    author = {Kaplan, Jared and McCandlish, Sam and Henighan, Tom and Brown, Tom B. and Chess, Benjamin and Child, Rewon and Gray, Scott and Radford, Alec and Wu, Jeffrey and Amodei, Dario},
    month = jan,
    year = {2020},
    note = {arXiv:2001.08361 [cs]},
}

@inproceedings{lin_learning_2023,
    title = {Learning {Vision}-and-{Language} {Navigation} from {YouTube} {Videos}},
    url = {https://openaccess.thecvf.com/content/ICCV2023/html/Lin_Learning_Vision-and-Language_Navigation_from_YouTube_Videos_ICCV_2023_paper.html},
    booktitle = {IEEE/CVF Conference on Computer Vision and Pattern Recognition},
    language = {en},
    urldate = {2025-12-12},
    author = {Lin, Kunyang and Chen, Peihao and Huang, Diwei and Li, Thomas H. and Tan, Mingkui and Gan, Chuang},
    year = {2023},
    pages = {8317--8326},
}

@phdthesis{rosenfeld_scaling_2021,
    type = {PhD Thesis},
    title = {Scaling {Laws} for {Deep} {Learning}},
    copyright = {In Copyright - Educational Use Permitted},
    url = {https://dspace.mit.edu/handle/1721.1/139897},
    language = {en},
    urldate = {2026-03-23},
    school = {Massachusetts Institute of Technology},
    author = {Rosenfeld, Jonathan S.},
    month = sep,
    year = {2021},
}

@inproceedings{suomela_synthetic_2025,
    title = {Synthetic vs. {Real} {Training} {Data} for {Visual} {Navigation}},
    url = {https://openreview.net/forum?id=ukFkk1BtpW},
    language = {en},
    urldate = {2026-03-23},
    author = {Suomela, Lauri and Arachchige, Sasanka Kuruppu and Torres, German F. and Edelman, Harry and Kämäräinen, Joni-Kristian},
    month = sep,
    year = {2025},
    booktitle = {Robot Data Workshop at CoRL 2025},
}

@inproceedings{black__2025,
    title = {\(\pi_0\): {A} {Vision}-{Language}-{Action} {Flow} {Model} for {General} {Robot} {Control}},
    volume = {21},
    isbn = {979-8-9902848-1-4},
    shorttitle = {\(\pi_0\)},
    url = {https://www.roboticsproceedings.org/rss21/p010.html},
    urldate = {2026-03-23},
    author = {Black, Kevin and Brown, Noah and Driess, Danny and Esmail, Adnan and Equi, Michael Robert and Finn, Chelsea and Fusai, Niccolo and Groom, Lachy and Hausman, Karol and Ichter, Brian and Jakubczak, Szymon and Jones, Tim and Ke, Liyiming and Levine, Sergey and Li-Bell, Adrian and Mothukuri, Mohith and Nair, Suraj and Pertsch, Karl and Shi, Lucy Xiaoyang and Smith, Laura and Tanner, James and Vuong, Quan and Walling, Anna and Wang, Haohuan and Zhilinsky, Ury},
    month = jun,
    year = {2025},
    booktitle = {Robotics: Science and Systems XXI},
}

@article{black_real-time_2025,
    title = {Real-{Time} {Execution} of {Action} {Chunking} {Flow} {Policies}},
    volume = {38},
    url = {https://openreview.net/forum?id=UkR2zO5uww},
    language = {en},
    urldate = {2026-03-23},
    journal = {Advances in Neural Information Processing Systems},
    author = {Black, Kevin and Galliker, Manuel Y. and Levine, Sergey},
    month = oct,
    year = {2025},
}


\end{document}